\documentclass[publish,JRM,paper]{jaciiiarticle}
\usepackage[pdftex]{graphicx,graphics, color}
\usepackage{epsfig}  
\usepackage{slashbox}
\usepackage{float}
\usepackage{amsmath}
\usepackage{amssymb}
\usepackage{mwe}
\usepackage{siunitx}
\usepackage{multirow}
\usepackage{cite}
\usepackage{subcaption}
\usepackage{caption}
\usepackage{bm}
\usepackage{color}
\usepackage{txfonts}
\usepackage{textcomp}
\usepackage{mathcomp}
\usepackage{url}

\SetVolumeNumber{Vol.0 No.0, 200x}
\SetArticleID{}
\begin{document}
\pagestyle{jaciiistyle}

\title{GREEMA : Proposal and Experimental Verification of Growing Robot by Eating Environmental MAterial for Landslide Disaster}
\author{Yusuke Tsunoda, Yuya Sato, and Koichi Osuka}
\address{2-1, Yamadaoka, Suita, Osaka 565-0871, JAPAN \\
    E-mail: y.tsunoda@mech.eng.osaka-u.ac.jp, sato.y@dsc.mech.eng.osaka-u.ac.jp, osuka@mech.eng.osaka-u.ac.jp}
\markboth{Yusuke Tsunoda}{Growing Robot by Eating Environmental Material}
\dates{00/00/00}{00/00/00}
\maketitle

\begin{abstract}
    \noindent 
    In areas that are inaccessible to humans, such as the lunar surface and landslide sites, there is a need for multiple autonomous mobile robot systems that can replace human workers.
    In particular, at landslide sites such as river channel blockages, robots are required to remove water and sediment from the site as soon as possible.
    Conventionally, several construction machines have been deployed to the site for civil engineering work.
    However, because of the large size and weight of conventional construction equipment, it is difficult to move multiple units of construction equipment to the site, resulting in significant transportation costs and time.
    To solve such problems, this study proposes a novel growing robot by eating environmental material called GREEMA, which is lightweight and compact during transportation, but can function by eating on environmental materials once it arrives at the site.
    GREEMA actively takes in environmental materials such as water and sediment, uses them as its structure, and removes them by moving itself.
    In this paper, we developed and experimentally verified two types of GREEMAs.
    First, we developed a fin-type swimming robot that passively takes water into its body using a water-absorbing polymer and forms a body to express its swimming function.
    Second, we constructed an arm-type robot that eats soil to increase the rigidity of its body.
    We discuss the results of these two experiments from the viewpoint of Explicit-Implicit control and describe the design theory of GREEMA.
\end{abstract}

\begin{keywords}
    mobile robot, growing robot, open design approach, implicit control, landslide dam
\end{keywords}

\section{Introduction}
\begin{figure}[t]
    \begin{minipage}[]{\columnwidth}
     \centering
     \includegraphics[width=0.7\columnwidth]{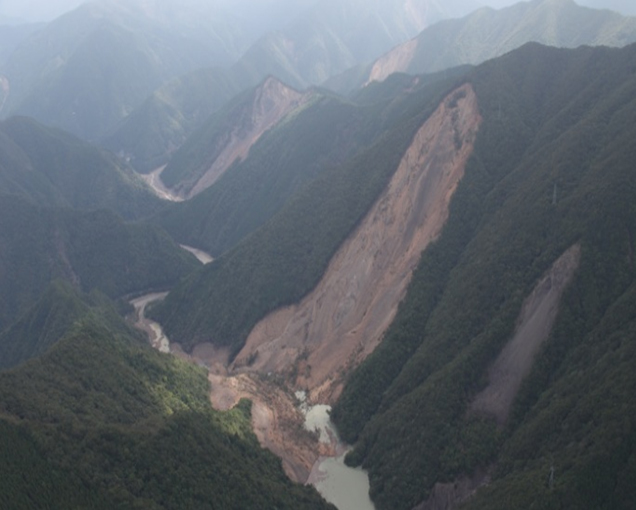}
     \caption{Natural dams in Totsukawa River in Nara prefecture (2011)[a]}
     \label{fig:kadou}
    \end{minipage}
\end{figure}

River channel blockage is one of the most common landslide disasters in Japan.
In this disaster, a river is dammed by sediments due to landslides caused by heavy rainfall, forming a natural dam, as shown in Fig.~\ref{fig:kadou}.
Then, as the water level of the river rises, the deposited sediment breaks through the dam, causing a debris flow\cite{akazawa2014numerical,zheng2021recent,fan2021recent}.
Mudslides cause extensive damage to homes in downstream areas.
Therefore, at the site of a river channel blockage, it is necessary to remove water and sediment from the site as soon as possible to prevent the spread of damage\cite{Wataru_SAKURAI2019,Hiroaki_SUGAWARA2019}.

Conventional emergency restoration work for this disaster has so far mainly used construction equipment and drainage pumping systems for civil engineering work.
For example, backhoes and dump trucks are used to excavate and transport earth and soil, and pump hose systems are used to suck up and remove water from dammed lakes.
However, because of the large size and mass of the machines conventionally used, it is very difficult to transport them to the site in the first place, and even if it were possible, it would take a significant amount of time for the construction equipment to reach the site.
In fact, in the conventional approach, large work equipment must be disassembled and transported by helicopter on multiple trips.
In addition, workers must assemble the construction equipment once it arrives at the site, which is extremely dangerous in the unpredictable and changing environment of a landslide site.
In addition to river channel blockage, it is also impossible to transport multiple robots by space shuttles when they are large and heavy, such as when mining or performing infrastructure constructions on the moon's surface\cite{otsu2013terrain,6497341}.

\begin{figure}[t]
    \begin{minipage}[]{1\columnwidth}
    \centering
    \includegraphics[width=0.7\columnwidth]{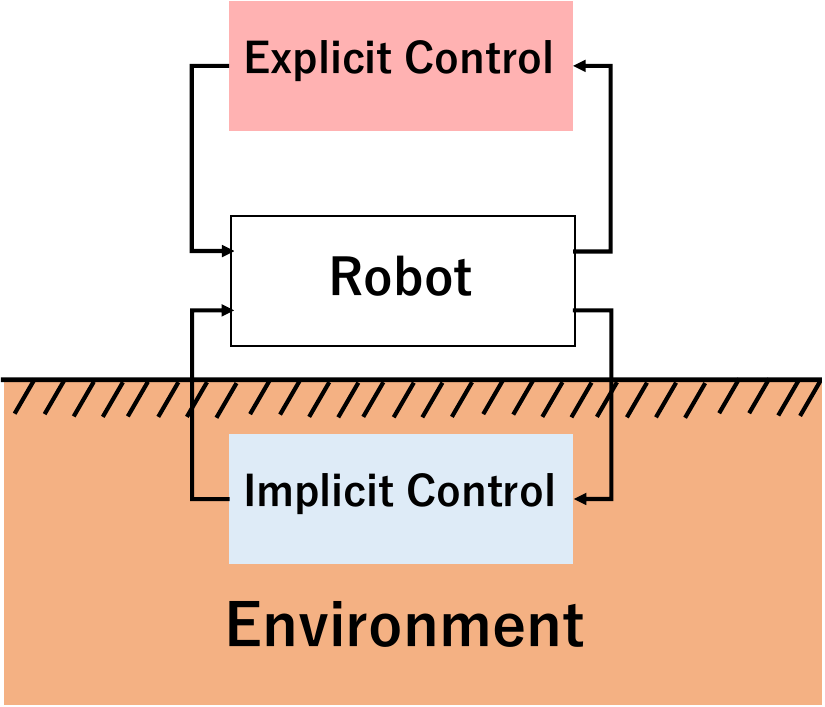}
    \caption{Conceptual diagram of implicit control and explicit control}
    \label{fig:previous}
    \end{minipage}
\end{figure}

To solve these problems, we aim to develop a completely new robot specialized for landslide response that is small and lightweight\cite{open_design}.
In this project, we have designed a robot based on the concept of ``implicit control''\cite{osuka2010dual,201028_491,5653968}.
Implicit control is a robot design concept proposed by Osuka et al..
The concept suggests that we don't fight head-on against the natural environment, which is an unlimited environment, but rather achieves its goal by making good use of the interaction between the environment and the robot's body.
Fig.~\ref{fig:previous} shows the conceptual diagram of implicit control.
In a harsh environment such as a landslide site, it is important to devise the structure and dynamics of the robot so that not only the control given to the robot (explicit control) but also the interaction between the robot body and the environment can be considered a kind of control (implicit control).

\begin{figure}[t]
    \begin{minipage}[]{1\columnwidth}
    \centering
    \includegraphics[width=0.8\columnwidth]{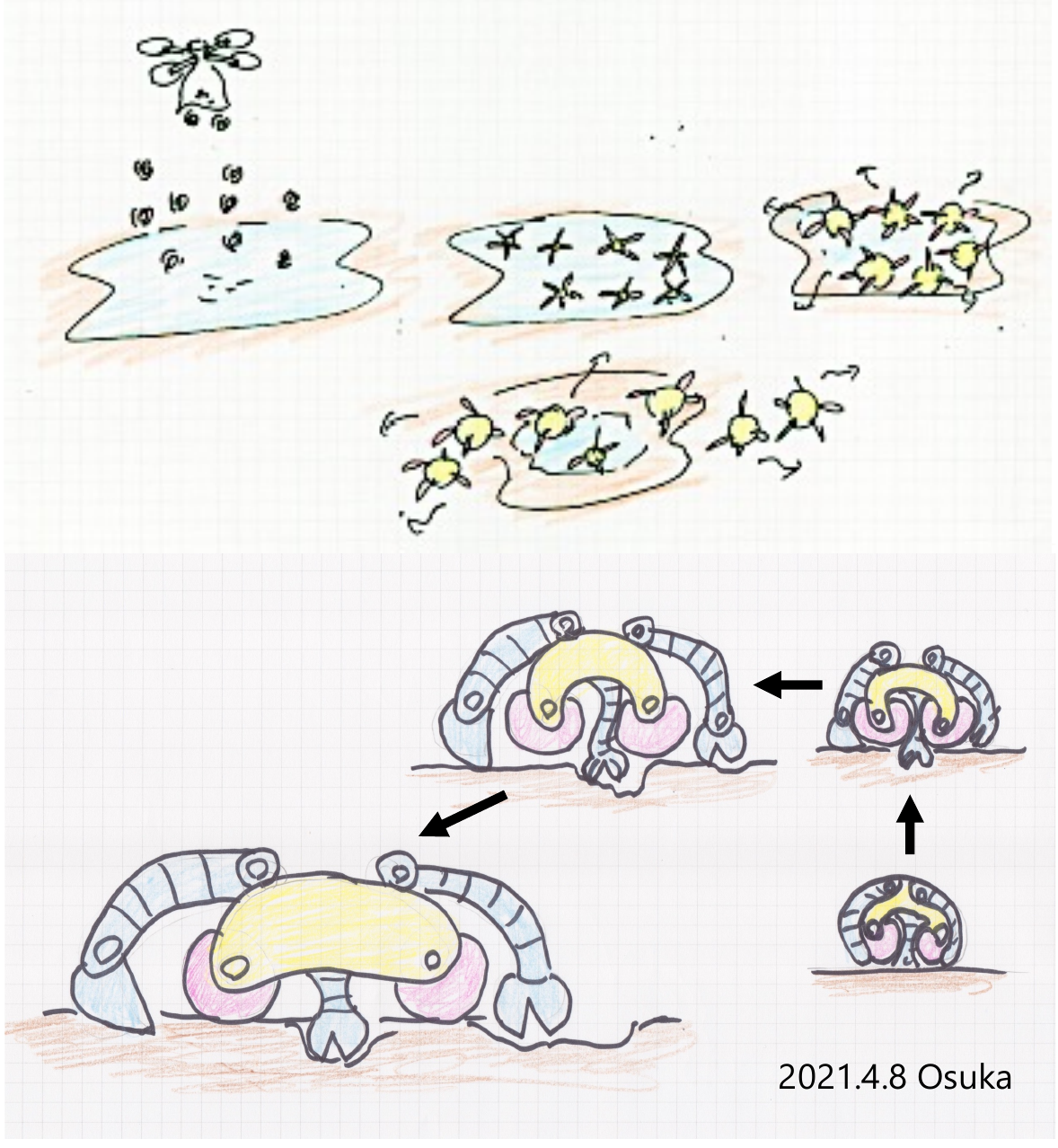}
    \caption{Concept image of GREEMA (Growing Robot by Eating Environmental MAterial)}
    \label{fig:greema}
    \end{minipage}
\end{figure}

\begin{figure}[t]
    \begin{minipage}[]{1\columnwidth}
        \centering
        \includegraphics[width=0.7\columnwidth]{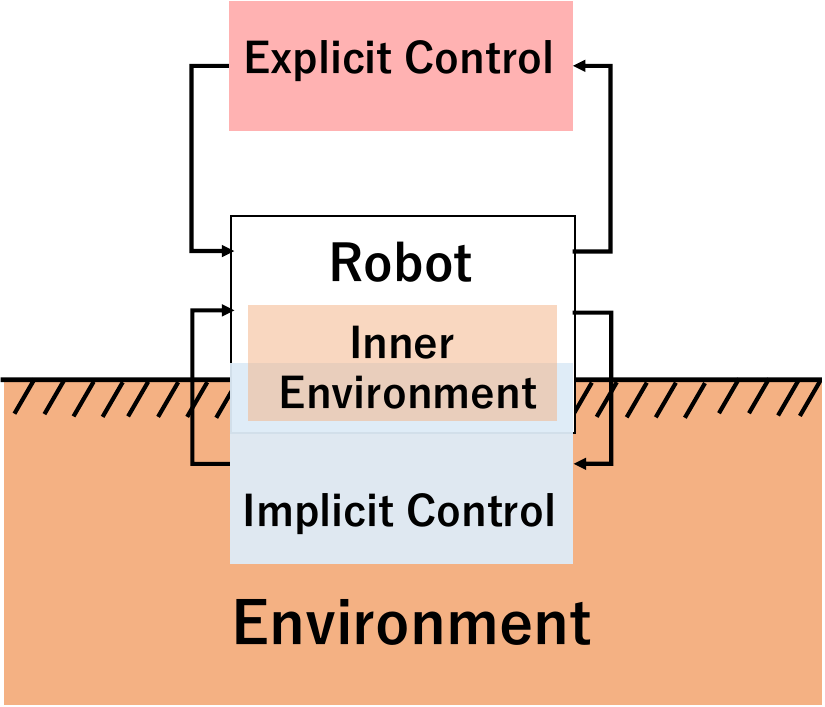}
        \caption{Proposed robot system from the viewpoint of implicit-explicit control}
        \label{fig:propose}
    \end{minipage}
\end{figure}

Based on this idea, we propose a robot called ``GREEMA'' (Growing Robot by Eating Environmental MAterial), which autonomously takes in environmental materials such as local sediment and water into its body and removes them by moving itself.
A conceptual diagram of the idea of GREEMA is shown in Fig.~\ref{fig:greema}.
GREEMA's body is small and light during transportation, but after arriving at the disaster site, it grows by acquiring its own weight, rigidity, and body parts by taking in sediment and water in the robot body.
After that, the robot aims to remove the sediment and water by moving itself.
GREEMA has two advantages: (1) the robot can be transported to the site in large quantities at a time, and (2) it can be expected to perform other tasks (e.g., digging for soil or transporting fallen trees) after acquiring and growing environmental materials.
A conceptual diagram of GREEMA's control system from the viewpoint of implicit control is shown in Fig.~\ref{fig:propose}.
As shown in Fig.~\ref{fig:propose}, GREEMA can be interpreted as taking in the environment (Inner environment) by eating local environmental material and modifying its own body so that its interaction with the environment can be considered implicit control.
Similar research to this study has focused on the physicality of robots, and there is research on robots that achieve control objectives by dramatically changing their body structure and dynamics\cite{origami,belke2017mori,sihite2023multi,5229953}.
This method is different from the proposed method in this study, in which the body is modified by eating environmental objects in the natural environment.
Moreover, some studies use environmental objects such as air\cite{vine,ilievski2011soft}, water\cite{ando2020fire,takahashi2020retraction}, sand\cite{jamming_arm}, ice\cite{ice}, and tree branches\cite{kinoeda} as part of the robot's hardware or as a driving method for actuators, but GREEMA is designed to autonomously take environmental objects into the body at the site of a landslide disaster and actively utilize them for physical deformation.
In addition, there is research on robots that eat materials in the outside world, focusing on the predatory behavior of living organisms\cite{wilkinson2000gastrobots,ieropoulos2003artificial,ieropoulos2005ecobot,philamore2015row}.
While these robots aim to obtain energy by eating the materials, the GREEMA aims to remove sediment and water by taking them into the robot's body and moving them around.

In this paper, as the first step of the development of GREEMA, we develop GREEMA robots that feed on two different kinds of environmental materials and express their functions.
The first is a fin-type swimming robot expressing a swimming function by taking water from its environment and forming a body.
We adopt superabsorbent polymers(SAP, water-absorbent polymers) to passively absorb water\cite{Kohei_UENO202322-00251}, and test whether they can passively acquire a flexible body and develop a swimming function by absorbing water.
The second is an arm-type robot that increases its rigidity by eating soil and taking them into its body.
We propose a mechanism in which a torus-shaped cloth is drawn into the inside and outside of the hose, which serves as the arm, to draw soil into the hose by internal friction and verify the validity of the proposed mechanism through an actual robot experiment.
Furthermore, we show that the rigidity of the machine changes depending on the water content ratio of the captured soil.

This paper is organized as follows.
Section 2 shows the first verification of our proposed GREEMA, the swimming robot with fins that can swim by taking in water and forming a body.
Section 3 presents the second validation, the arm-type robot that gains rigidity by feeding on soil and taking them into its body.
In Section 4, we discuss the design theory of GREEMA through these two experimental results based on implicit control.
Finally, Section 5 concludes the paper.


\section{Experimental Verification (1): The Swimming Robot by Eating Water}
To verify the validity of the proposed GREEMA, we developed the fish-type robot that forms its body using water as an environmental material and expresses a swimming function.
One application of the water-eating GREEMA is to quickly remove water that accumulates at the site of a landslide disaster, such as a blocked river channel, by absorbing water and moving the robot.
GREEMA can also form a body in the presence of water and thus has potential applications such as generating a swarm of robots to explore the seafloor or removing oil leaks from wrecked ships by absorbing the water.

\begin{figure}[t]
    \begin{minipage}[]{1\columnwidth}
        \centering
        \includegraphics[width=0.9\columnwidth]{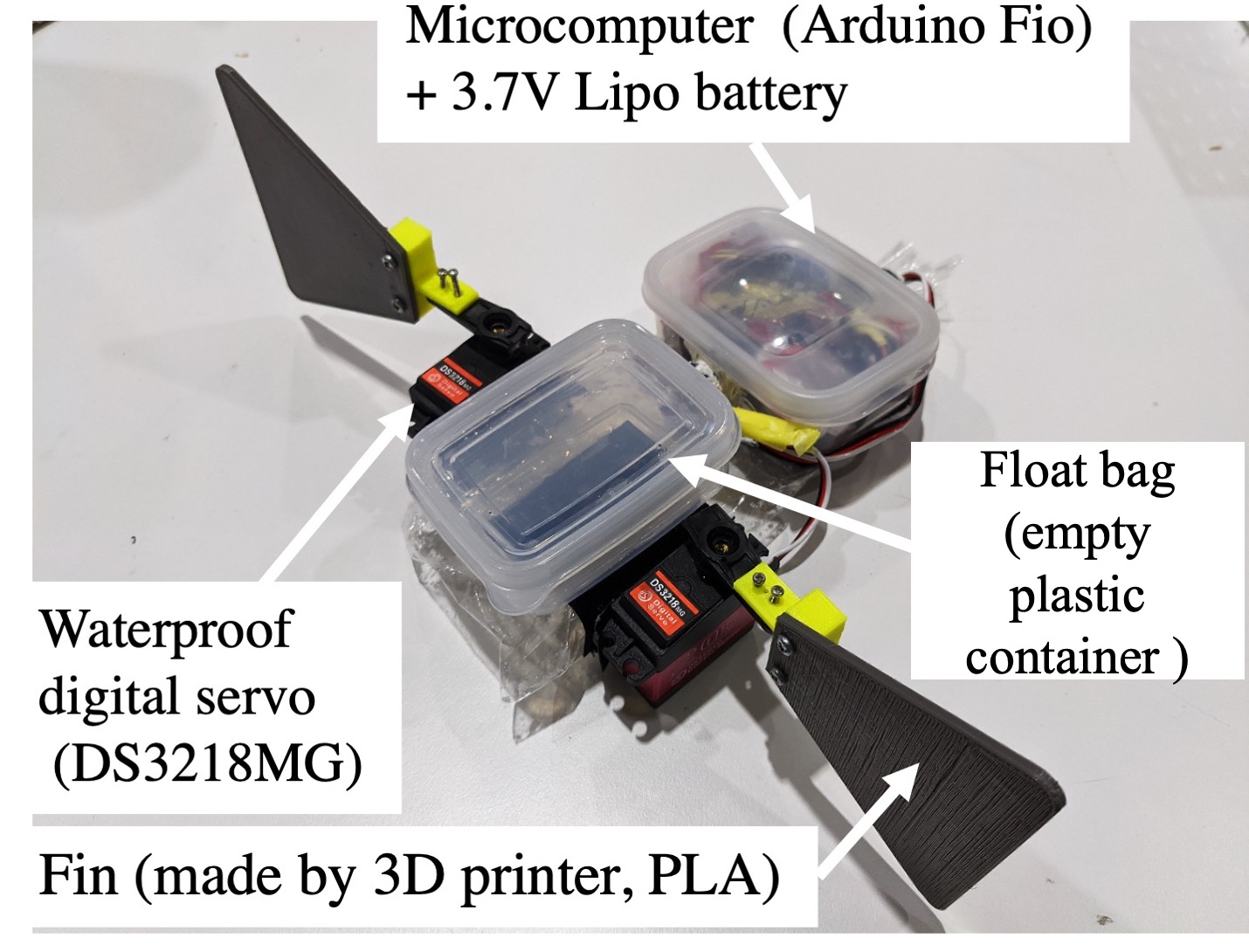}
        \subcaption{Top view}
        \label{fig:top}
    \end{minipage}
    \begin{minipage}[]{1\columnwidth}
        \centering
        \includegraphics[width=0.9\columnwidth]{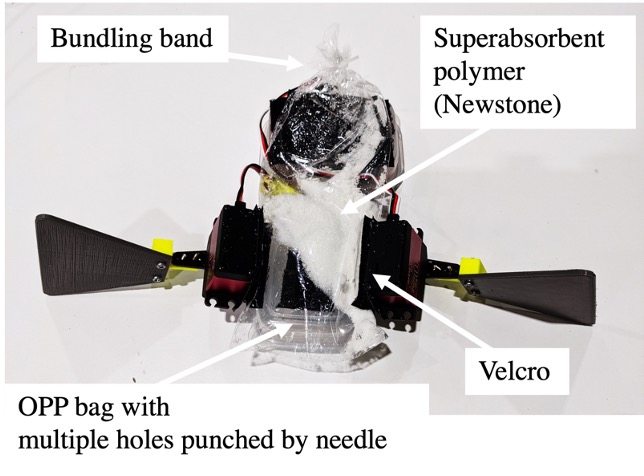}
        \subcaption{Bottom view}
        \label{fig:bottom}
    \end{minipage}
    \caption{Developed growing robot growing by eating water:``Mizu-Kurai''}
    \label{fig:fish_robot}
\end{figure}
\begin{figure}[t]
    \begin{minipage}[]{\columnwidth}
     \centering
     \includegraphics[width=0.8\columnwidth]{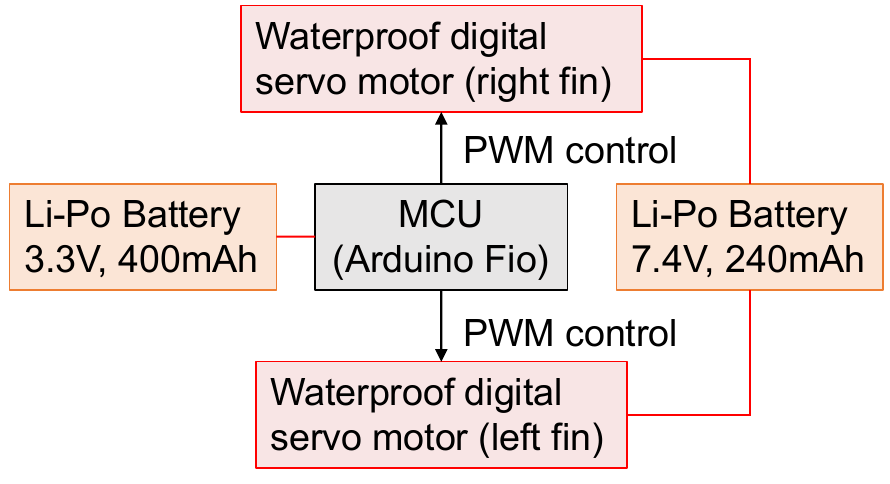}
     \caption{Control system of the ``Mizu-Kurai'' }
     \label{fig:system_aqua}
    \end{minipage}
\end{figure}
\begin{figure}[t]
        \begin{minipage}[]{1\columnwidth}
        \centering
        \includegraphics[width=0.6\columnwidth]{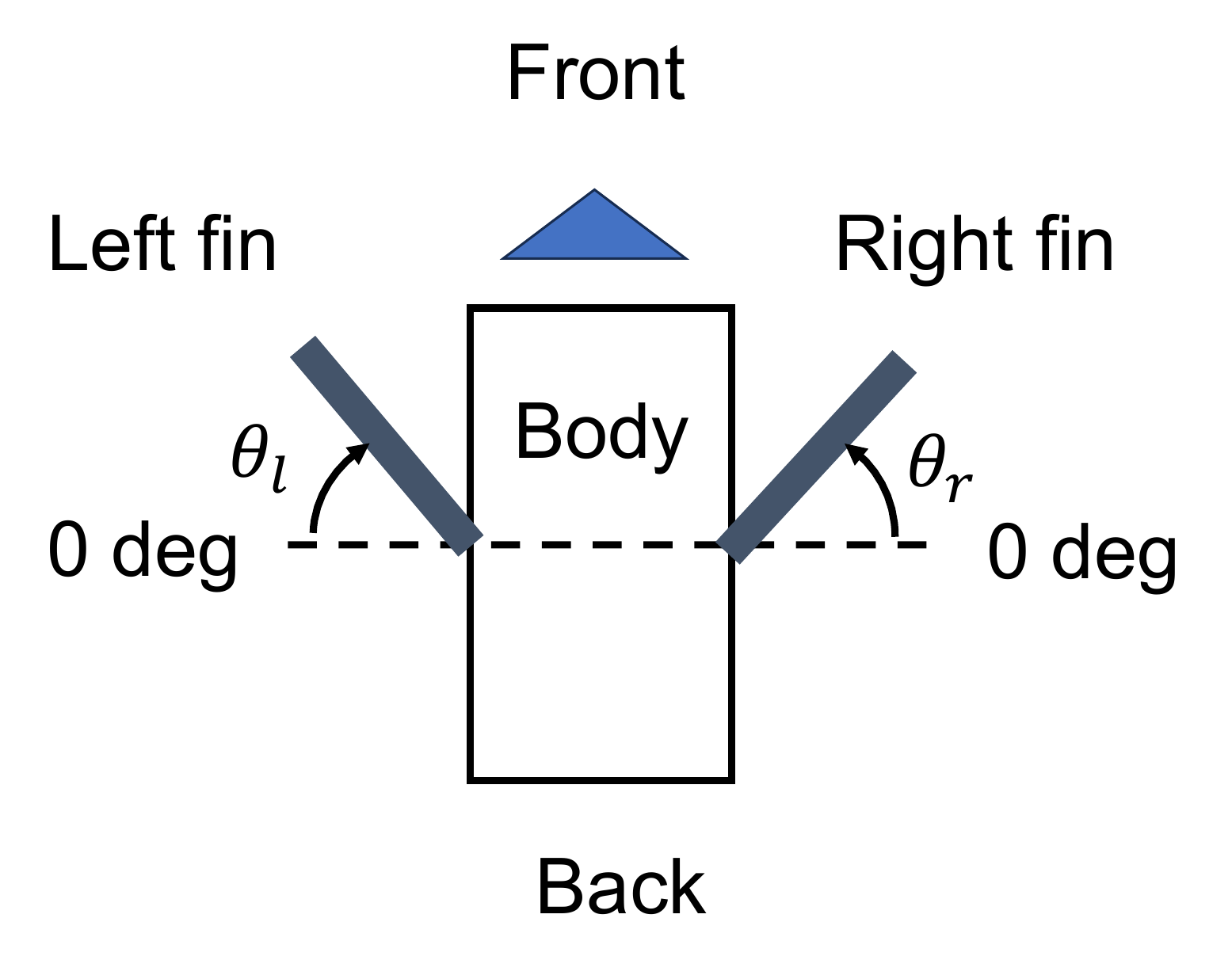}
        \subcaption{Definition of fin angle $\theta_r$ and $\theta_l$}
        \label{fig:model}
        \end{minipage}
       \begin{minipage}[]{1\columnwidth}
        \centering
        \includegraphics[width=\columnwidth]{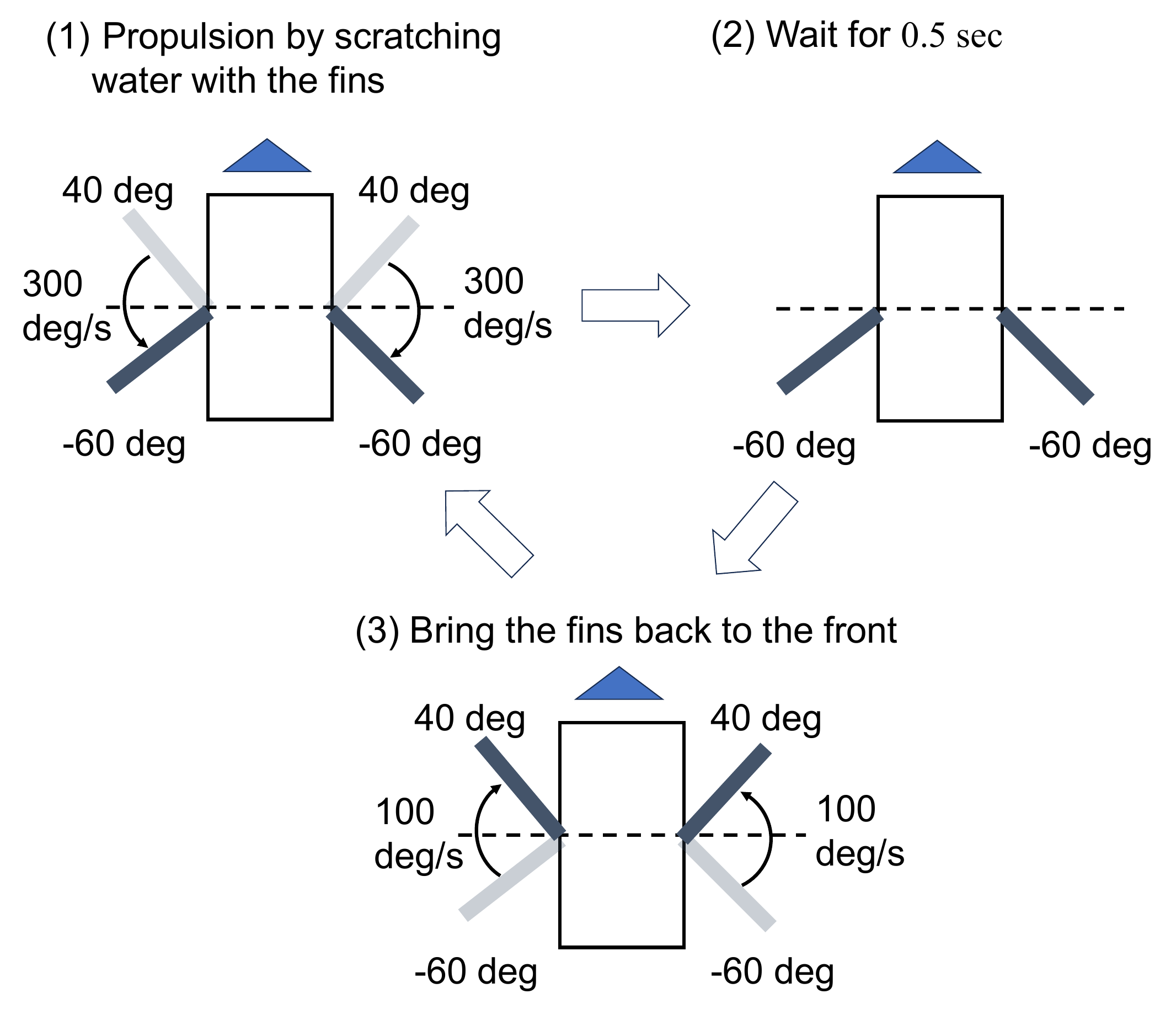}
        \subcaption{Algorithm of the fin control}
        \label{fig:argo}
    \end{minipage}
    \caption{Design of the controller of the proposed robot}
\end{figure}
\begin{figure}[t]
    \begin{minipage}[]{1\columnwidth}
        \centering
        \includegraphics[width=0.85\columnwidth]{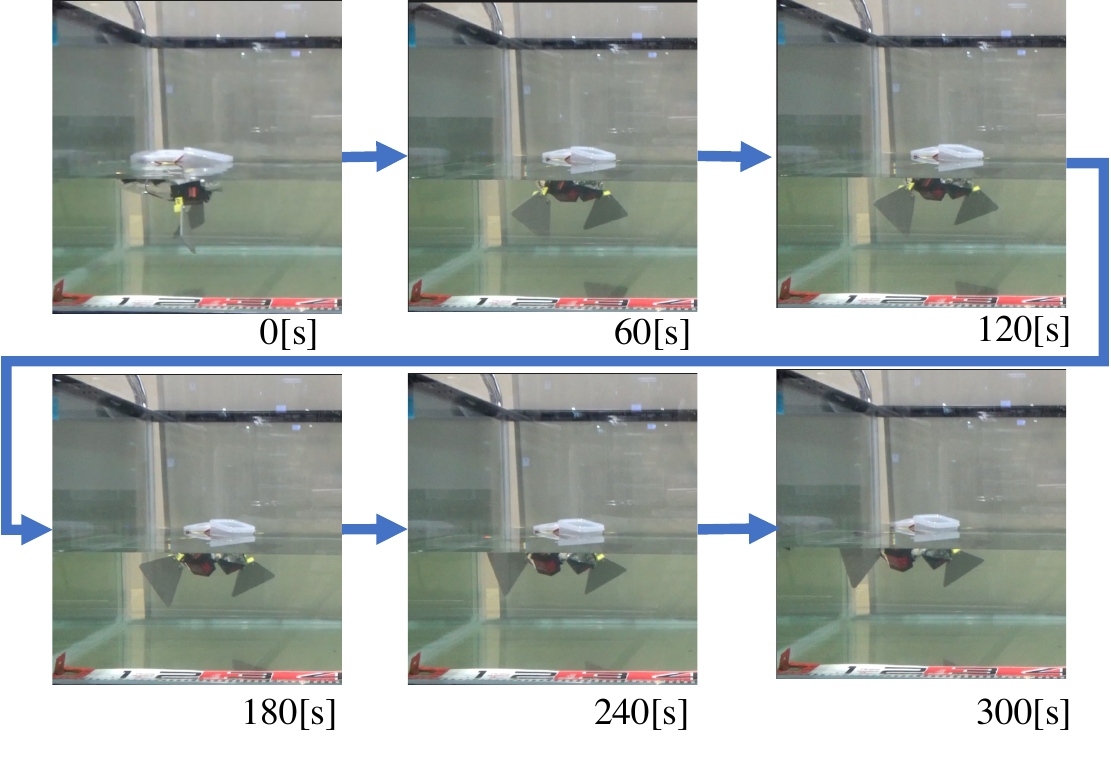}
        \subcaption{Snapshots of the process of body formation through water absorption}
        \label{fig:fin}
    \end{minipage}
    \begin{minipage}[]{1\columnwidth}
        \centering
        \includegraphics[width=0.8\columnwidth]{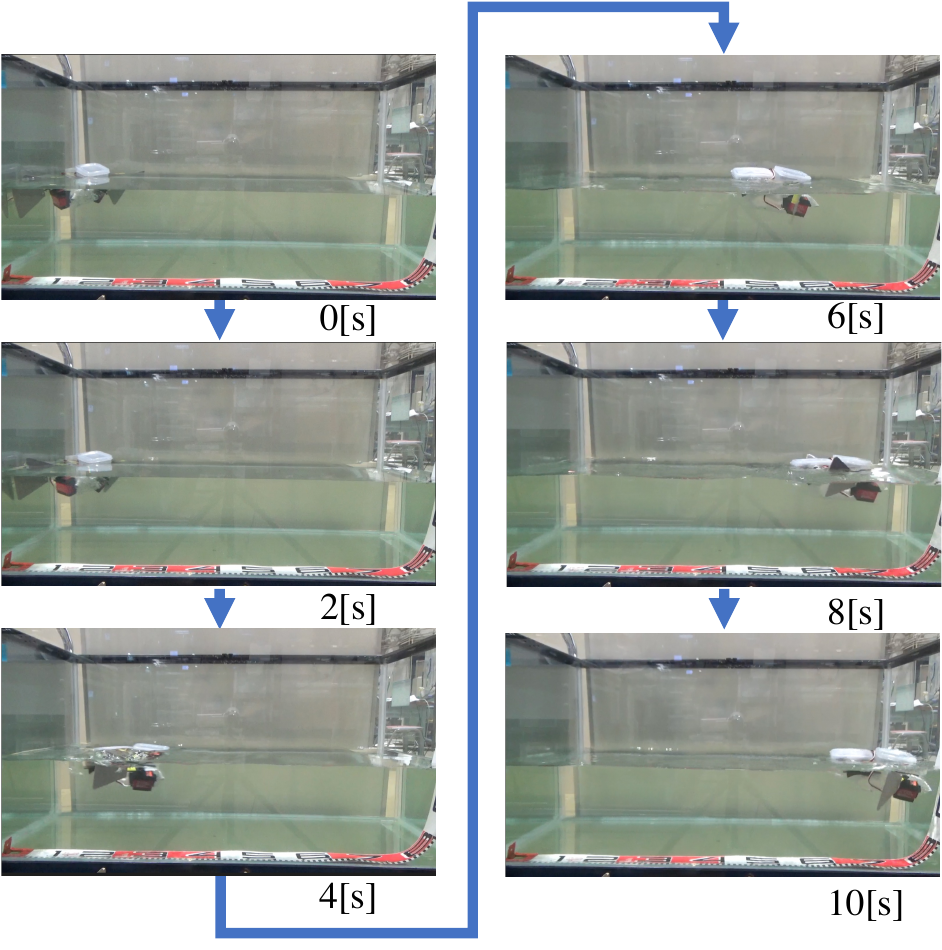}
        \subcaption{Snapshots of the movement of the swiming robot with superabsorbent polymers (experiment (1))}
        \label{fig:SAP}
    \end{minipage}
    \caption{Results of the swimming experiment of the proposed robot}
    \label{fig:success}
\end{figure}
\begin{figure}[t]
    \begin{minipage}[]{1\columnwidth}
        \centering
        \includegraphics[width=0.8\columnwidth]{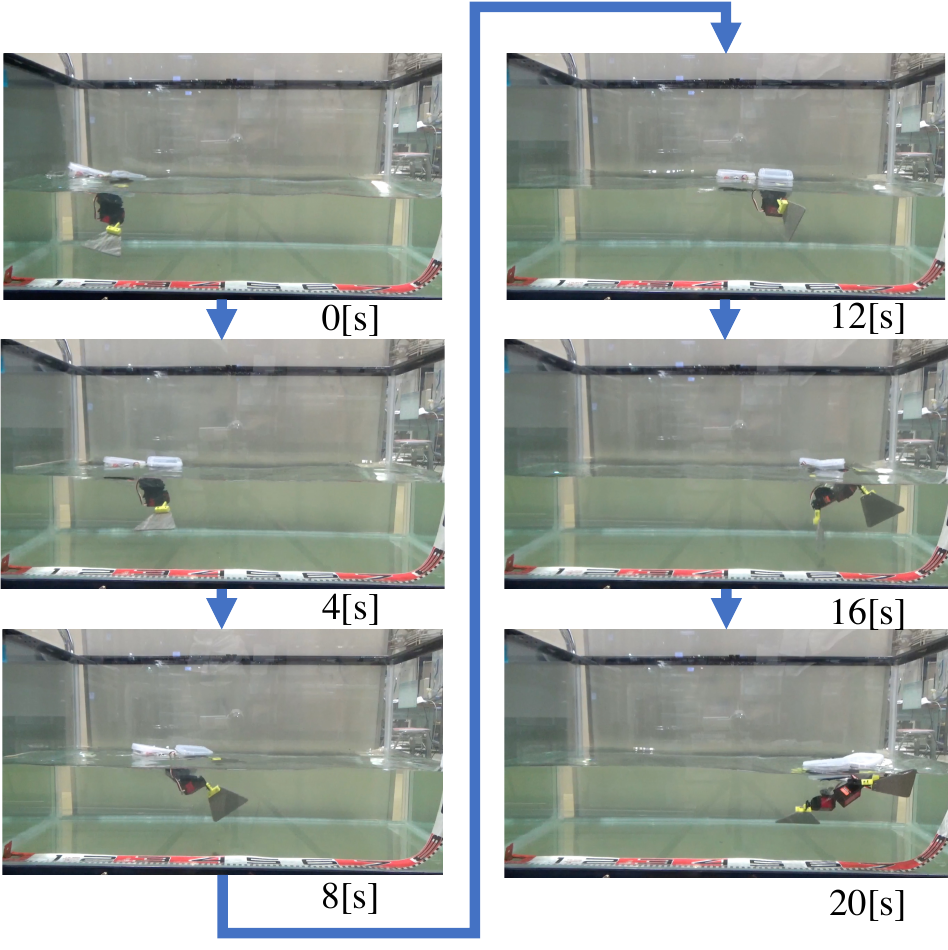}
        \caption{Snapshots of the movement of the swiming robot without superabsorbent polymers (experiment (2))}
        \label{fig:non_SAP}
    \end{minipage}
\end{figure}

\subsection{Developed Proposed Robot}
The appearance and dimensions of the developed robot named ``Mizu-Kurai'' are shown in Fig.~\ref{fig:fish_robot}.
In this paper, we adopt a body formation using a SAP so that the robot can passively absorb water and form a flexible body.
The robot's body is an OPP bag containing 15 g of superabsorbent polymer (Newstone).
The bag is perforated with a needle to allow water to enter from the outside.
One waterproof digital servo motor (DS3218MG, Goolsky) is attached to each side of the fuselage via velcro, and trapezoidal fins (made of PLA) formed by a 3D printer are attached to the tips of these motors.
The microcontroller board is Arduino Fio, and the servo is driven by PWM control.
The battery and electronic circuits are placed in a small plastic container and installed behind the robot.
To balance the buoyancy between the front and rear of the robot, an empty plastic container is installed in front of the robot as a float.
The maximum dimensions of the robot are 23 cm (length), 30 cm (width), and 10 cm (height), and the total weight is 318 g, including the SAP.
In this paper, considering the volume of the OPP bag in the body and the water absorption rate of the SAP, 15 g of the SAP is sealed in the bag in advance, and the robot is controlled to move its fins and swim 15 minutes after landing on water.

Figure.~\ref{fig:system_aqua} shows the system configuration of the water-eating robot.
The microcontroller is used to PWM-control two waterproof digital servo motors that move the left and right fins.
A 3.3V, 400mAh Li-Po battery was used to drive the microcontroller,and a 7.4V, 240mAh Li-Po battery was used to drive the servo motors.

\subsection{Robot Controller of Mizu-Kurai}
Next, we describe the control of the fins.
As shown in Fig.~\ref{fig:model}, we define that the angle of the robot's right fin at a time $t~(t \geq 0)$ from the robot's true side (positive clockwise) is $\theta_r[t] (-90 \leq \theta_r \leq 90 )$ and $\theta_l[t] (-90 \leq \theta_l \leq 90)$ for the angle (positive counterclockwise) of the left fin from the side of the robot.
In this case, the following feed-forward control is applied to the fin angle $\theta_r[t]$ and $\theta_l[t]$, respectively, as shown in Fig.~\ref{fig:argo}.
\begin{enumerate}
\item[(1)] The left and right flippers are moved at an angular velocity of $300$ deg/s from $\theta_r[t] = \theta_l[t] = 40$deg to $\theta_r[t] = \theta_l[t] = -60$deg.
This action causes the robot to move forward by paddling the water.
\item[(2)] The robot maintains $\theta_r[t] = \theta_l[t] = -60$deg for $0.5$s. The reason for this is to take advantage of the propulsive force obtained by plucking water backward with the fins.
\item[(3)] Each fin is moved at an angular velocity of $100$ deg/s until the angle of each fin reaches $\theta_r[t] = \theta_l[t] = 40$deg. This is a movement to move the fins back forward to account for the drag force on the fins in the water.
\item[(4)] Return to step (1) and repeat steps (1)-(3).
\end{enumerate}

\subsection{Experimental Results and Discussion}
\begin{figure}[t]
    \begin{minipage}[]{1\columnwidth}
        \centering
        \includegraphics[width=0.85\columnwidth]{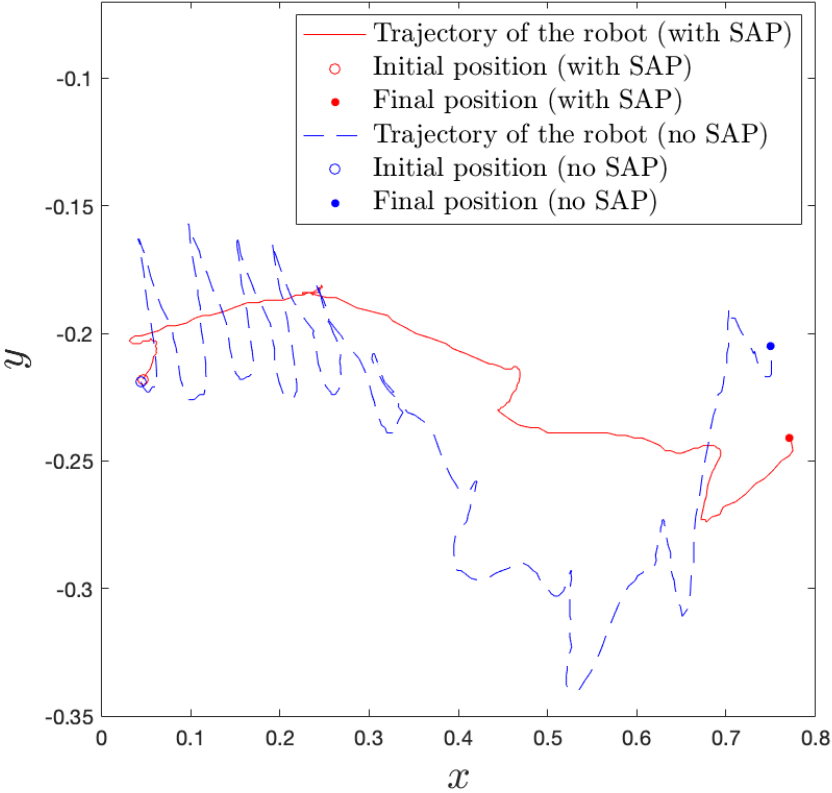}
        \caption{Trajectries of the robot in each case (red line: experiment (1), blue dashed line:experiment (2))}
        \label{fig:top_kiseki}
    \end{minipage}
\end{figure}
\begin{figure}[t]
    \begin{minipage}[]{1\columnwidth}
        \centering
        \includegraphics[width=0.85\columnwidth]{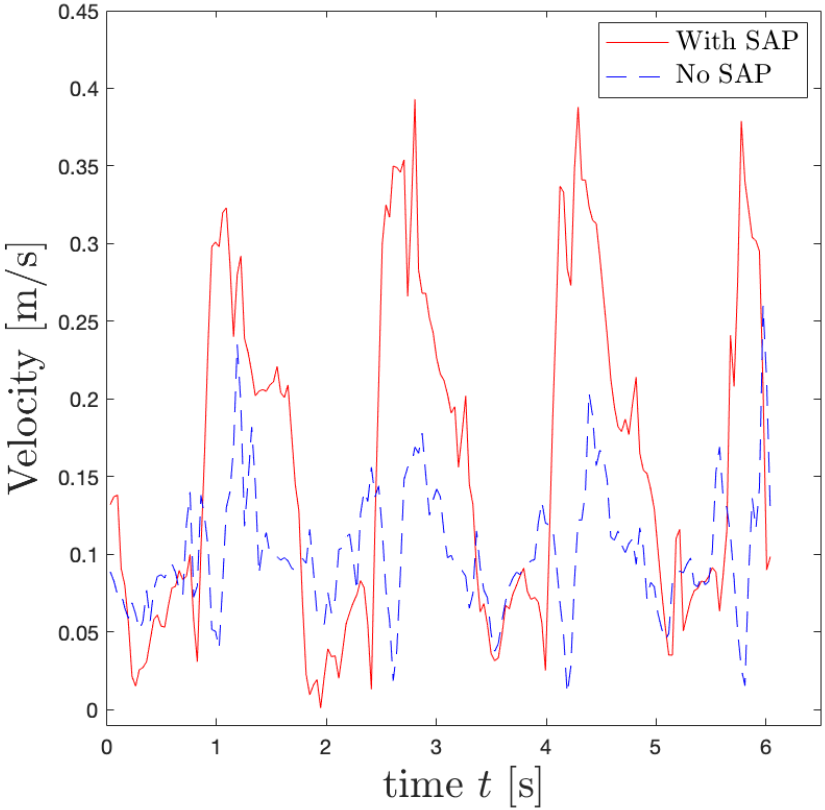}
        \caption{Velocity of the robot in each case (red line: experiment (1), blue dashed line:experiment (2))}
        \label{fig:top_v}
    \end{minipage}
\end{figure}
\begin{figure}[t]
    \begin{minipage}[]{1\columnwidth}
        \centering
        \includegraphics[width=0.85\columnwidth]{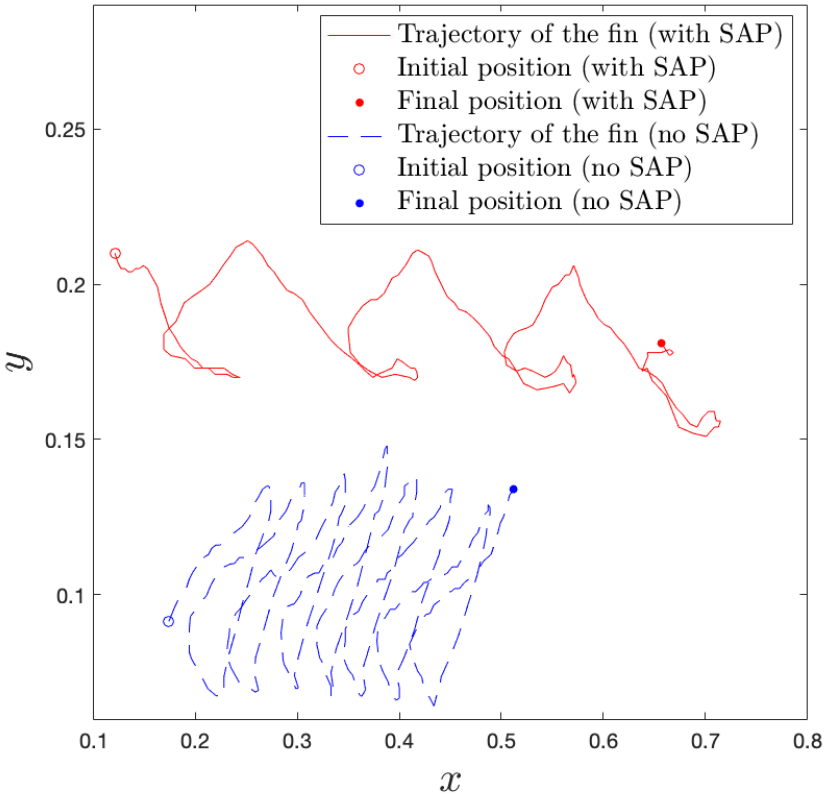}
        \caption{Trajectries of the robot's fin in each experiment (red line: experiment (1), blue dashed line:experiment (2))}
        \label{fig:side_kiseki}
    \end{minipage}
\end{figure}
\begin{figure}[t]
    \begin{minipage}[]{1\columnwidth}
        \centering
        \includegraphics[width=0.85\columnwidth]{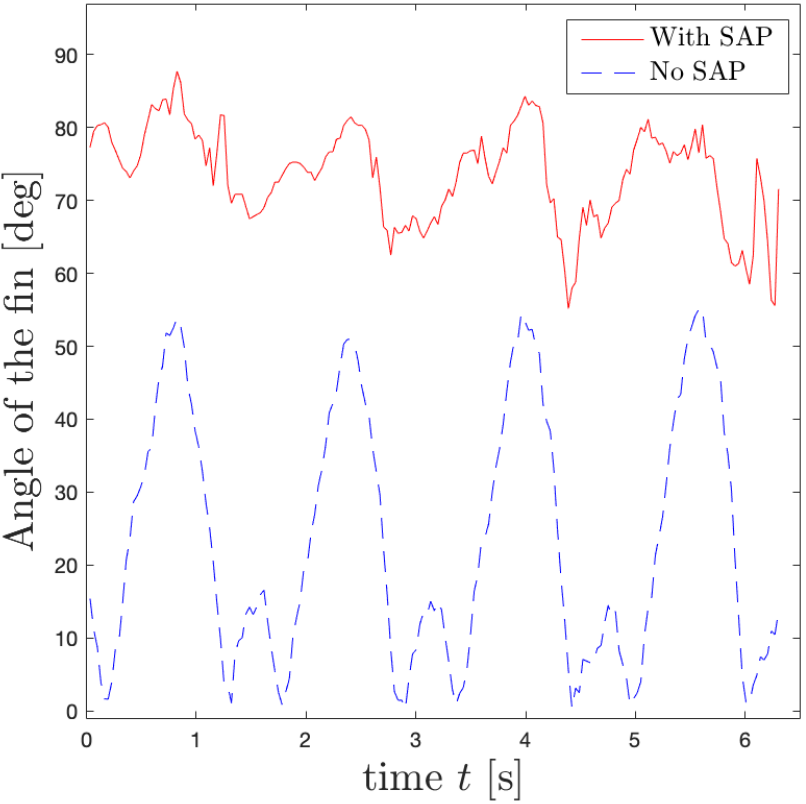}
        \caption{Degree angle of the robot's fin in each case (red line: experiment (1), blue dashed line:experiment (2))}
        \label{fig:hire_deg}
    \end{minipage}
\end{figure}

To confirm whether the robot can form a flexible body and swim by absorbing water with SAP, we conducted two types of experiments: (1) with SAP sealed inside the body and (2) without SAP sealed inside the body and compared the differences in behavior underwater.
The experimental environment was a water tank 90 cm long, 45 cm wide, and 45 cm high, filled to half its height (approximately 91 L) with tap water.

First, in case (1), the body growth process formed by the water absorption of the SAP is shown in Fig.~\ref{fig:fin}, and the swimming behavior that develops after that is shown in Fig.~\ref{fig:SAP}.
From Fig.~\ref{fig:fin}, the rigidity and volume of the body increase as the SAP passively absorb water, which causes the right and left fins to expand.
Fifteen minutes after landing on the water, the robot can be seen swimming, as shown in Fig.~\ref{fig:SAP}.
The robot had a flexible body made of SAP after absorbing water, and it passively acquired a body suitable for swimming through interaction with the environment.
The mass of the robot after the experiment was 643 g. 
If the density of water is 1.0 g/ml, the robot absorbs 325 ml of water in 15 minutes, and removing the robot is equivalent to removing this amount of water.
Although SAPs are inherently highly water absorbent (each 1 g of SAP absorbs 300 ml of water in 1 minute and 20 seconds), the robot showed less water absorption than the theoretical value.
This is because this experiment added a large amount of SAP to the OPP bag's volume to increase the body's rigidity after water absorption. 
As a result, the amount of water absorption is thought to have been reduced.

Next, a snapshot of the swimming experiment in case (2) is shown in Fig.~\ref{fig:non_SAP}.
As can be seen from Fig.~\ref{fig:non_SAP}, the robot's fins hang down because its torso remains deflated, resulting in insufficient propulsive force to allow it to swim properly.

We compare the swimming performance of the robots in each experiment with quantitative data and discuss the causes.
Fig.~\ref{fig:top_kiseki} shows the trajectory of the robot as seen from the top in each experiment.
This trajectory represents the center of the tapper for the float behind the robot and was created by image processing the experimental movie taken from the top.
The image processing software Tracker was used.
As shown in Fig.~\ref{fig:top_kiseki}, when the SAP is not added, the robot initially oscillates from side to side and its movement is unstable, but when the SAP is added, the robot swims stably.
Fig.~\ref{fig:top_v} shows the time variation of the robot's movement speed in each experiment, and Table~\ref{tab:ave_v_result} shows the time average of the robot's movement speed.
We calculate this movement speed using the movement trajectory data shown in Fig.~\ref{fig:top_kiseki}.
From Fig.~\ref{fig:top_v}, the robot with SAP has a higher moving speed than the robot without SAP.
The reason for this is that the SAP absorbs water, which increases the rigidity of the robot's torso, and the left and right fins are supported stably.
In addition, Fig.~\ref{fig:side_kiseki} shows the trajectory of the center of the long side of the robot's right fin in each experiment.
This trajectory was obtained by Tracker image processing of a swimming movie taken from the lateral direction.
The height of the water surface is around $y=0.2$.
Fig.~\ref{fig:side_kiseki} shows that the robot without SAP has drooping fins, so its fins are always in the water and oscillate up and down during swimming.
In contrast, the robot with the SAP moves its fins near the surface of the water.
Furthermore, the angle of the long side of the right fin to the $x$-axis (the direction in which the robot is moving) is shown in Fig.~\ref{fig:hire_deg} in each experiment.
As shown in Fig.~\ref{fig:hire_deg}, the angle of the robot's fins without the SAP changes significantly and periodically around $0\sim55$ deg relative to the robot's direction of motion, whereas the angle of the robot with the SAP remains around $60\sim85$ deg.
The time averages of the fin angles are shown in Table~\ref{tab:ave_deg_result}, and the robot with the SAP had its fins develop near perpendicular to the direction of motion.
\begin{table}[t]
    \caption{Result of the robot's average velocity in each experiment}
    \label{tab:ave_v_result}
    \centering
    \begin{tabular}{|c|}
        \hline
        Case (1): with superabsorbent polymers \\\hline
        0.158 [m/s]  \\\hline
        Case (2): no superabsorbent polymers \\\hline
        0.101 [m/s] \\\hline
    \end{tabular}
\end{table}
\begin{table}[t]
    \caption{Result of the average angle of the fin in each experiment (red line: experiment (1), blue dashed line:experiment (2))}
    \label{tab:ave_deg_result}
    \centering
    \begin{tabular}{|c|}
        \hline
        Case (1): with superabsorbent polymers \\\hline
        73.607 [deg]  \\\hline
        Case (2): no superabsorbent polymers \\\hline
        24.087 [deg] \\\hline
    \end{tabular}
\end{table}
\section{Experimental Verification (2): The Arm Robot by Eating Soil}
\begin{figure}[t]
    \centering 
    \includegraphics[width=0.85\columnwidth]{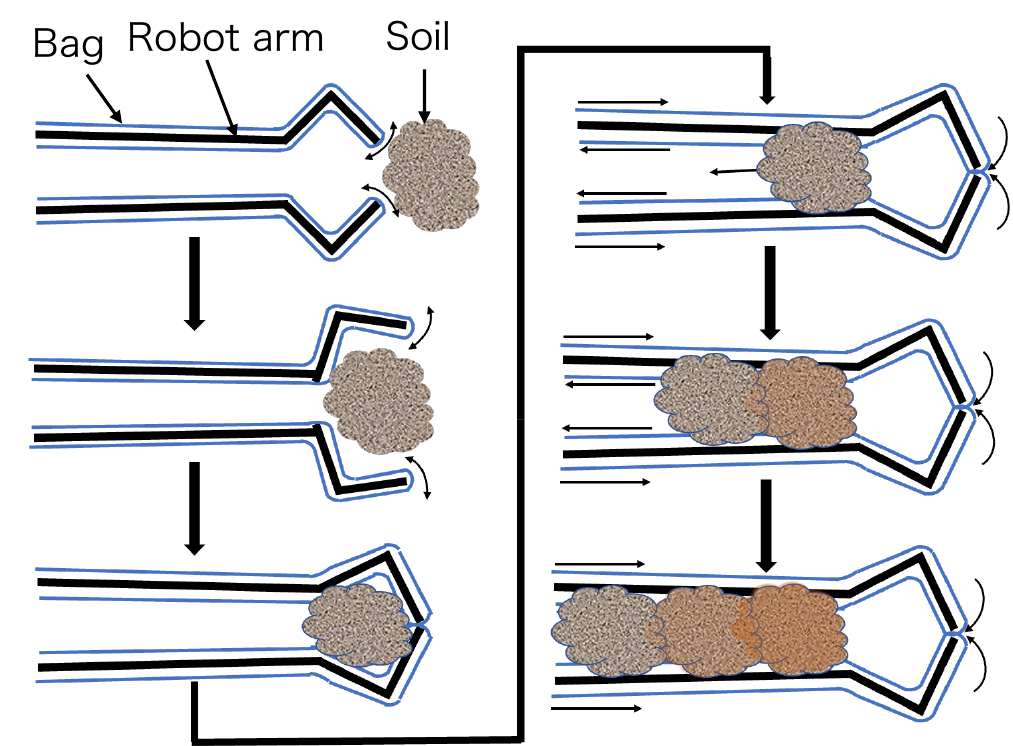}
    \caption{Collection and transportation of soil by proposed system}
    \label{fig:sikumi}
\end{figure}
\begin{figure}[t]
    \begin{minipage}[t]{\columnwidth}
        \centering
        \includegraphics[width=0.95\columnwidth]{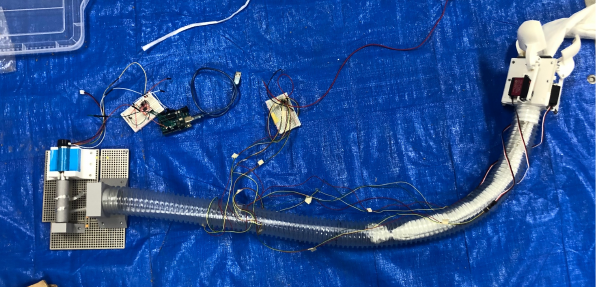}
        \subcaption{Overview}
        \label{fig:overview}
    \end{minipage}
    \begin{minipage}[b]{0.54\columnwidth}
        \centering
        \includegraphics[width=\columnwidth]{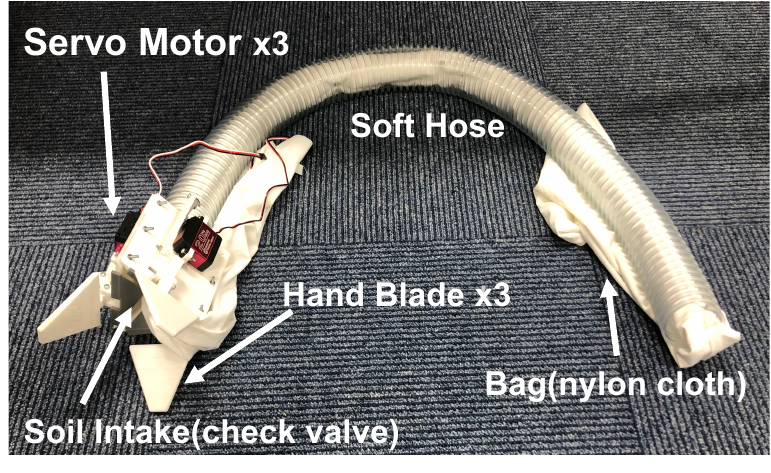}
        \subcaption{Gripper unit and hose body}
        \label{fig:sentan}
    \end{minipage}
    \begin{minipage}[b]{0.45\columnwidth}
        \centering
        \includegraphics[width=\columnwidth]{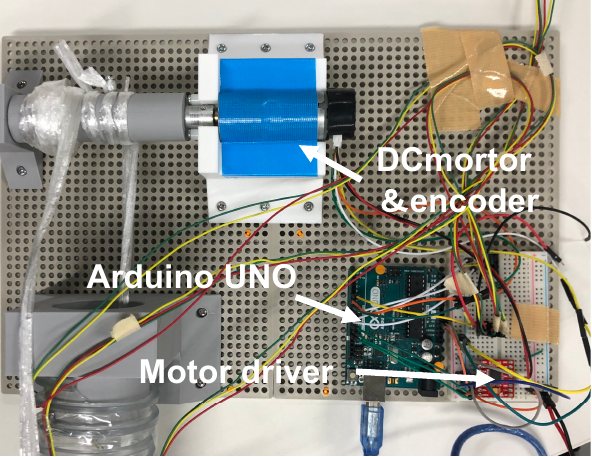}
        \subcaption{Bag winding mechanism}
        \label{fig:makitori}
    \end{minipage}
    \caption{Developed growing robot growing by eating :``Tsuchi-Kurai''}
    \label{fig:tsuchikurai}
\end{figure}
\begin{figure}[t]
    \begin{minipage}[b]{0.49\columnwidth}
        \centering 
        \includegraphics[width=0.7\columnwidth]{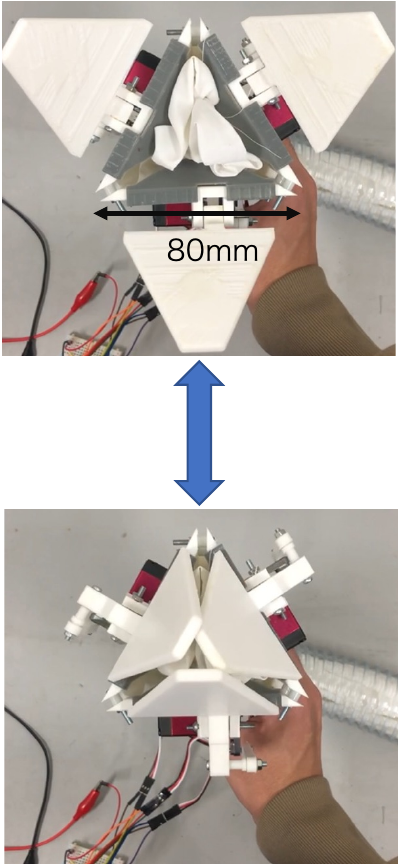}
        \subcaption{gripper}
        \label{fig:gripper_naked}
    \end{minipage}
    \begin{minipage}[b]{0.5\columnwidth}
        \centering 
        \includegraphics[width=0.7\columnwidth]{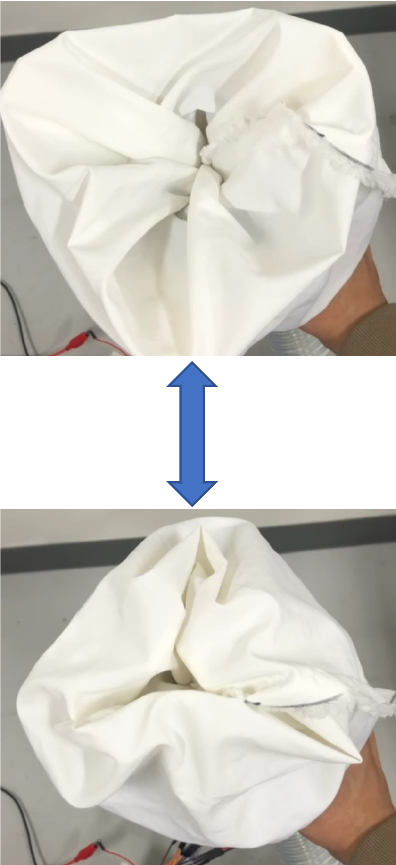}
        \subcaption{gripper covered with a bag}
        \label{fig:gripper_hukuro}
    \end{minipage}
    \caption{Opening and closing action of the gripper}
    \label{fig:gripper}
\end{figure}
\begin{figure}[t]
    \centering 
    \includegraphics[width=0.85\columnwidth]{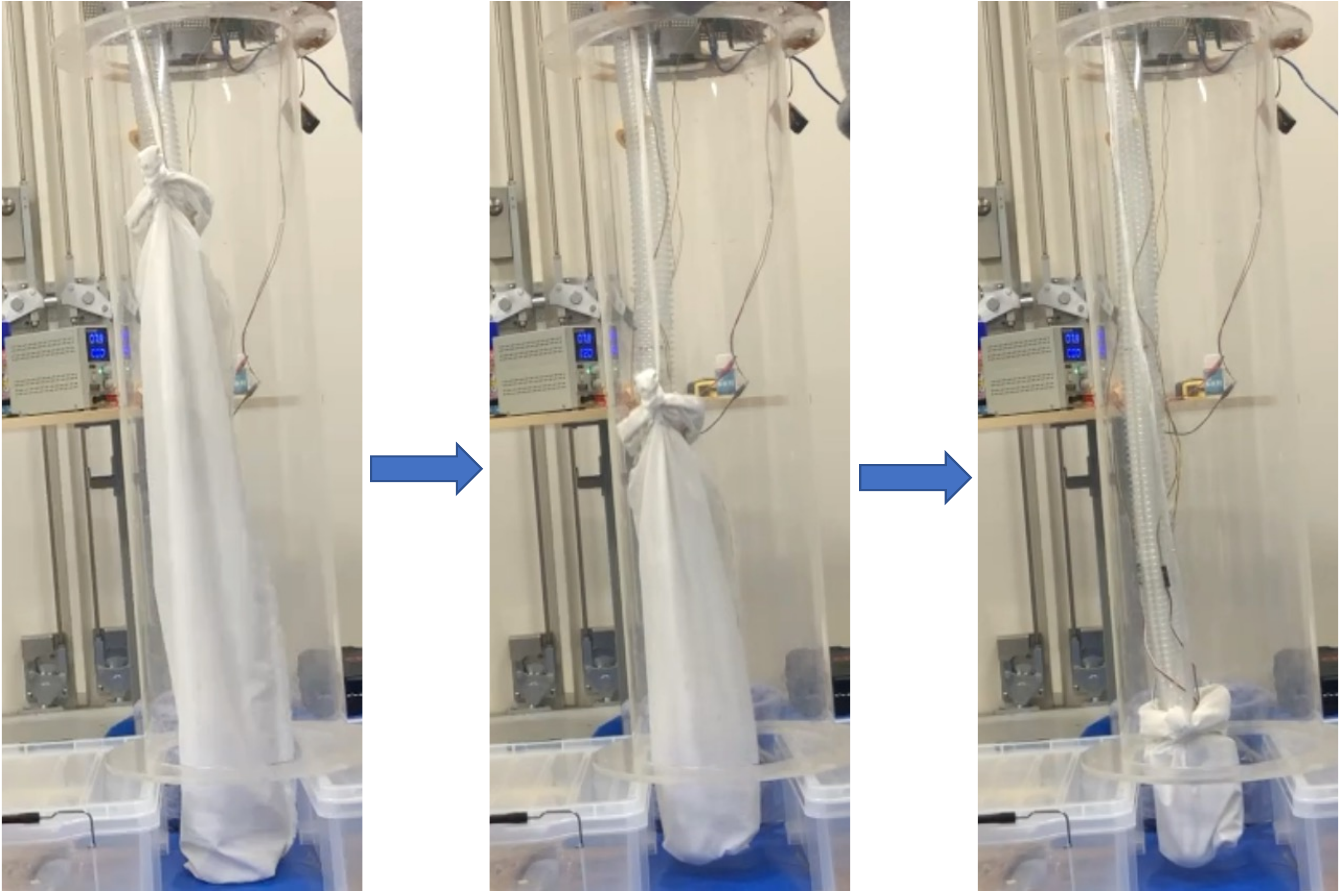}
    \caption{Bag retraction mechanism by wire winding}
    \label{fig:hikikomi}
\end{figure}
\begin{figure}[t]
    \begin{minipage}[]{\columnwidth}
     \centering
     \includegraphics[width=\columnwidth]{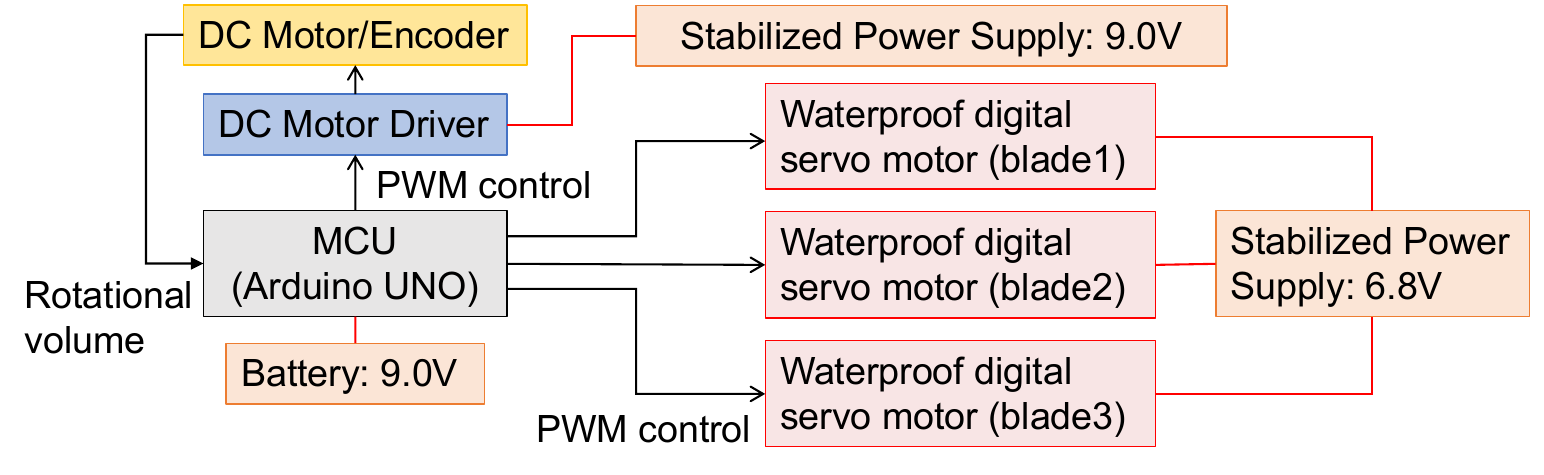}
     \caption{Control system of the ``Tsuchi-Kurai''}
     \label{fig:system_tsuchi}
    \end{minipage}
\end{figure}
\begin{figure}[t]
    \centering
    \includegraphics[width=0.7\columnwidth]{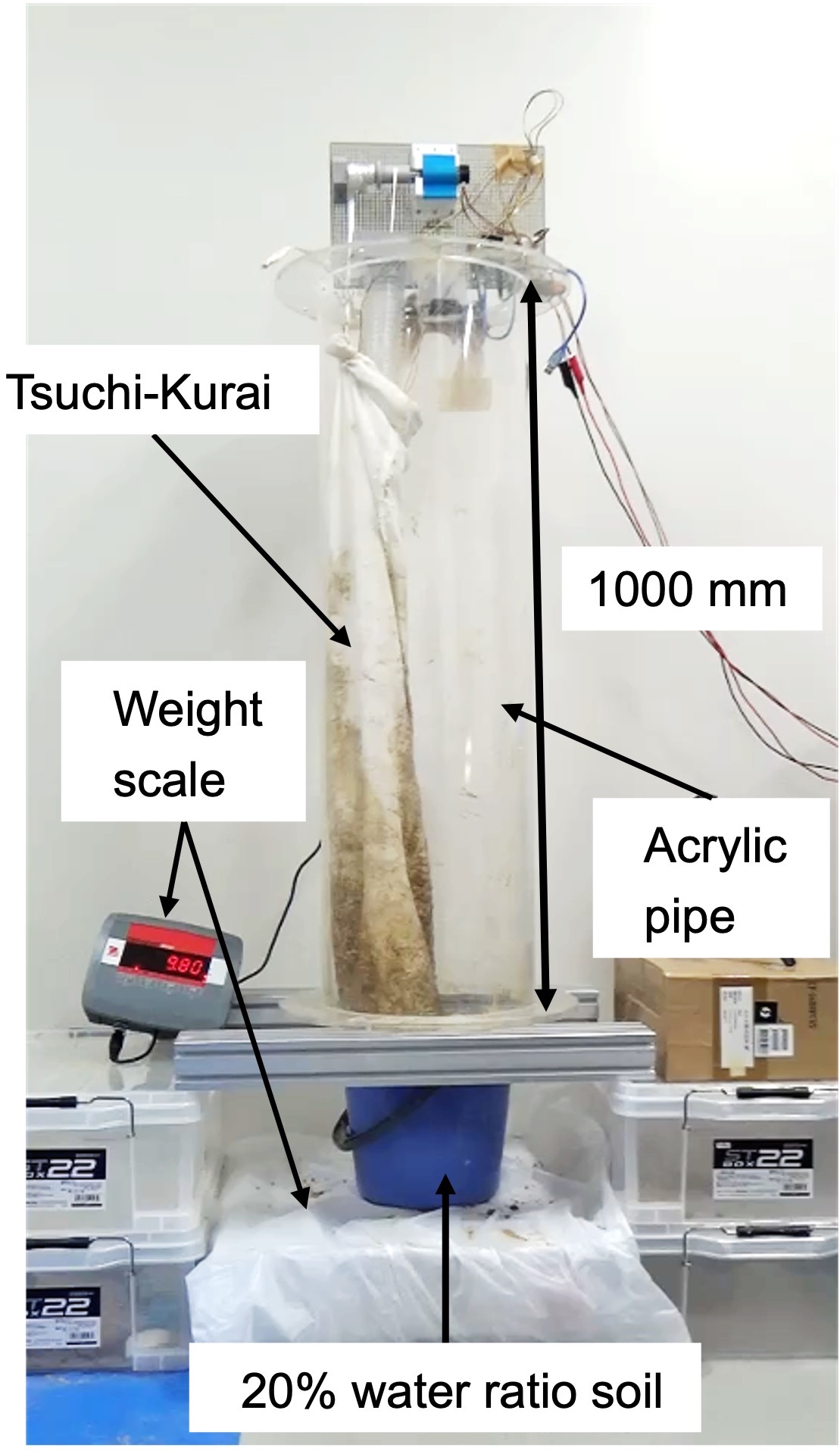}
    \caption{Soil collection experiment environment}\label{fig:torikomi_souchi}
\end{figure}
In this section, we describe the development of the arm-type robot that increases rigidity by eating soil and verifying the actual robot.

\subsection{Proposal of Mechanical Structure for Eating Soil}
We propose a robot hand mechanism encased in a bag-like structure to collect and transport soil, as shown in Fig.~\ref{fig:sikumi}.
This mechanism will collect and transport soil in the following steps.
\begin{enumerate}
    \item Robot hand with bags inside and outside
    \item Robot hand grabs soil
    \item Close the hand and wrap the soil inside the bag
    \item Retracts the bag and captures the soil
\end{enumerate}
This eating method does not require excavation of earth and sand deep into the ground and can be accomplished with low-power actuators.
In addition, the friction between the inside of the robot and the soil is slight because the robot moves with the soil wrapped in a bag-like structure.
Similar mechanisms include the torus-shaped bag as a gripper \cite{fujita2018jamming} and a similar mechanism for picking objects as if swallowing them \cite{torus_kaigai,torus_kaigai_2,swallowing}.
In addition, a mechanism in which a robot hand is attached to the tip of the torus gripper to support grasping has been proposed.
On the other hand, the proposed mechanism targets a fluid material called soil, has a tip-sealing function with a gripper covered by a bag, and aims to fill the acquired soil into the hose to make it its structure.

\subsection{Developed Proposed Robot}
Fig.~\ref{fig:tsuchikurai} shows the overview of the developed robot called ``Tsuchi-Kurai''.
As shown in Fig.~\ref{fig:overview}, the robot has a gripper part that eats soil, a flexible hose (duct hose made by MISUMI) that takes in soil, and a motor device that winds up the bag.
As shown in Fig.~\ref{fig:sentan}, the three blades of the gripper part open and close by using three servo motors.
The sediment intake is located in the center of the gripper section and is equipped with a check valve to prevent backflow of sediment.
By using a 3D printer, the gripper and three blades were made of PLA resin, and the check valve was made of TPU resin.
A nylon bag is provided to cover the inside of the hose and the gripper section (Fig.~\ref{fig:gripper}).
The robot can wind the bag with a DC motor with an encoder shown in Fig.~\ref{fig:makitori} and pull the bag part through the inside of the hose part as shown in Fig. ~\ref{fig:hikikomi}. 
The machine weighs 1.3 kg for the gripper and hose, 0.7 kg for the winder, 1.4 m in length, 60.2 mm in outer diameter and 50.6 mm in inner diameter for the hose, and a triangular cross-section of 80 mm per side for the gripper part.
Details of the microcontroller, motor driver, motor, and hose mounted on the robot are below.
\begin{itemize}
    \item A microcontroller (Arduino Uno Rev3)
    \item A Motor driver (TB6612FNG, SparkFun)
    \item A DC motor with encoder (GM25 OSOYOO)
    \item A Servo motor (DS3218MG from Goolsky) x3pic
    \item A Duct hose (HOSTOM-50 made by MISUMI) 1m
\end{itemize}

The system configuration diagram of the dirt eater is shown in Fig.~\ref{fig:system_tsuchi}.
Three waterproof digital servo motors are controlled by a microcontroller, and one DC motor with encoder is controlled by a DC motor driver for bag winding. 
The encoder sends the amount of bag winding rotation to the microcontroller.
A 9V battery was used to drive the microcontroller, a regulated power supply (6.8V) to drive the three servo motors, and a regulated power supply with 9.0V output to control the DC motor.

\subsection{Robot Controller of Tsuchi-Kurai}
Next, we describe the control algorithm of the Tsuchi-Kurai.
The robot operates according to the following algorithm.
\begin{enumerate}
    \item[(1)] The robot opens the three grippers until the blade angle of each gripper reaches $40$deg relative to the direction of the hose length, as shown in the upper figure of Fig.~\ref{fig:gripper_naked}.
    \item[(2)] The robot closes each gripper until its blade angle becomes $-15$deg relative to the hose length direction, as shown in the bottom figure of Fig.~\ref{fig:gripper_naked}. This eats up the sediment below the robot.
    \item[(3)] The DC motor is rotated to wind the wire $7$cm (for about $2$s), which pulls the bag inside the hose and pulls the sediment into the hose.
    \item[(4)] Return to (1) and repeat steps (1)-(3) until a total of $1$m of the bag is pulled in.
\end{enumerate}

\subsection{Soil Uptake Experiment}
We conducted the soil uptake experiment using the prototype machine developed.
Fig.~\ref{fig:torikomi_souchi} shows the appearance of the experimental apparatus.
As shown in Fig.~\ref{fig:torikomi_souchi}, the robot was set vertically downward, and a bucket containing mashatsuchi-soil with a moisture content of 20 \% was placed directly under the robot.
The water content of the soil was determined based on the assumption of a landslide disaster site caused by heavy rain.
The amount of soil acquired inside the machine was recorded by reading the change in the value of the weight scale under the bucket before and after the retrieval.

\begin{figure}[t]
    \centering
    \vspace{8mm}
    \includegraphics[width=0.7\columnwidth]{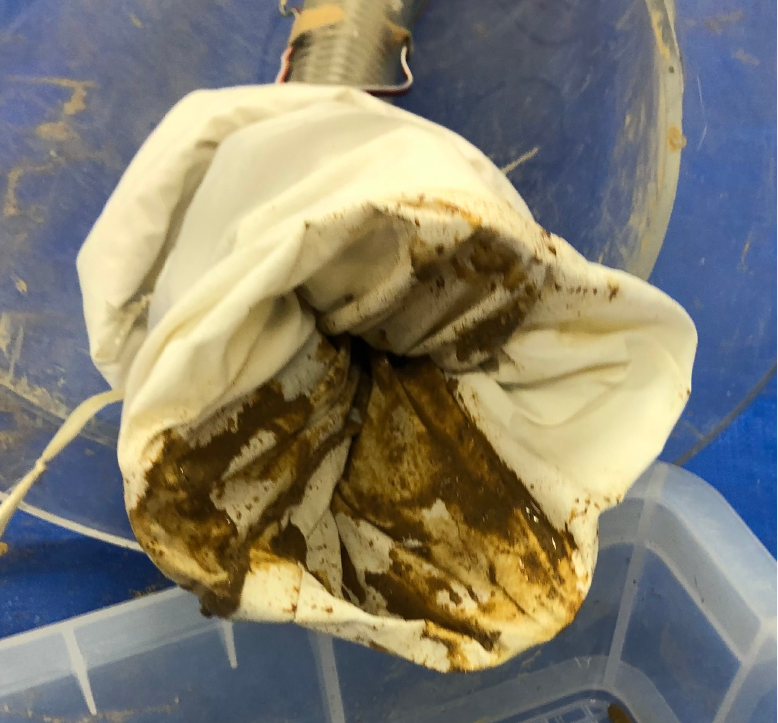}
    \caption{Inside the gripper after the experiment}\label{fig:jikkengo}
\end{figure}

Table \ref{tab:torikomi_result} shows the results of the amount of sediment acquired, the number of take-up motions, and the average amount of sediment acquired per take-up motion for each experiment.
The uptake operation is defined as the combination of opening and closing of the gripper and one bag take-up operation.
As shown in Table \ref{tab:torikomi_result}, the robot acquired up to $500$ g of sediment in one experiment and up to $31$ g of sediment per take-up operation.
This result indicates that the fluid sediment was captured and removed in an easily transportable object (the hose).
As shown in Fig.~(\ref{fig:jikkengo}), no sediment remains inside the gripper after the experiment, indicating that the acquired sediment is moved inside the hose section of the machine.
Based on these results, we conclude that the proposed mechanism can capture sediment and transport it within the robot.
Also, the reason for the difference in the number of take-up movements in each experiment is that in some cases, the bag was not pulled in even after the winder was turned because the bag was caught in the gripper.
The reason for the variation in the average amount of sediment acquired per uptake in each experiment may be that the robot operation was controlled by an external person in a feed-forward manner, which prevented the sediment from entering the gripper properly, or that the bag was retracted regardless of whether the gripper was able to grasp the sediment or not.
In future research, the robot should be equipped with a sensor to detect whether the robot itself is sufficiently grasping the sediment with the gripper and whether the sediment is being conveyed and filled into the hose by retracting the bag, and the design of the take-up operation by feedback control is necessary.
\begin{table}[t]
    \caption{Result of soil collection experiment}
    \label{tab:torikomi_result}
    \centering
    \begin{tabular}{|l|c|c|c|c|c|}
        \hline
        Experiments  & 1 & 2& 3 & 4 & 5 \\
        \hline \hline
        Soil acquisition[g] & 280 & 460 & 70  & 150 & 500 \\
        \hline
        Number of  & \multirow{2}{*}{20}  & \multirow{2}{*}{15}  & \multirow{2}{*}{14}  & \multirow{2}{*}{18}  & \multirow{2}{*}{19}  \\
        take-in operations & & & & & \\\hline
        Average amount of & & & & & \\
        soil acquired & \multirow{2}{*}{14} & \multirow{2}{*}{31}  & \multirow{2}{*}{5.0} & \multirow{2}{*}{8.3} & \multirow{2}{*}{26} \\
        per intake[g] & & & & & \\\hline
    \end{tabular}
\end{table}
\begin{figure}[t]
    \centering
    \includegraphics[width=0.75\columnwidth]{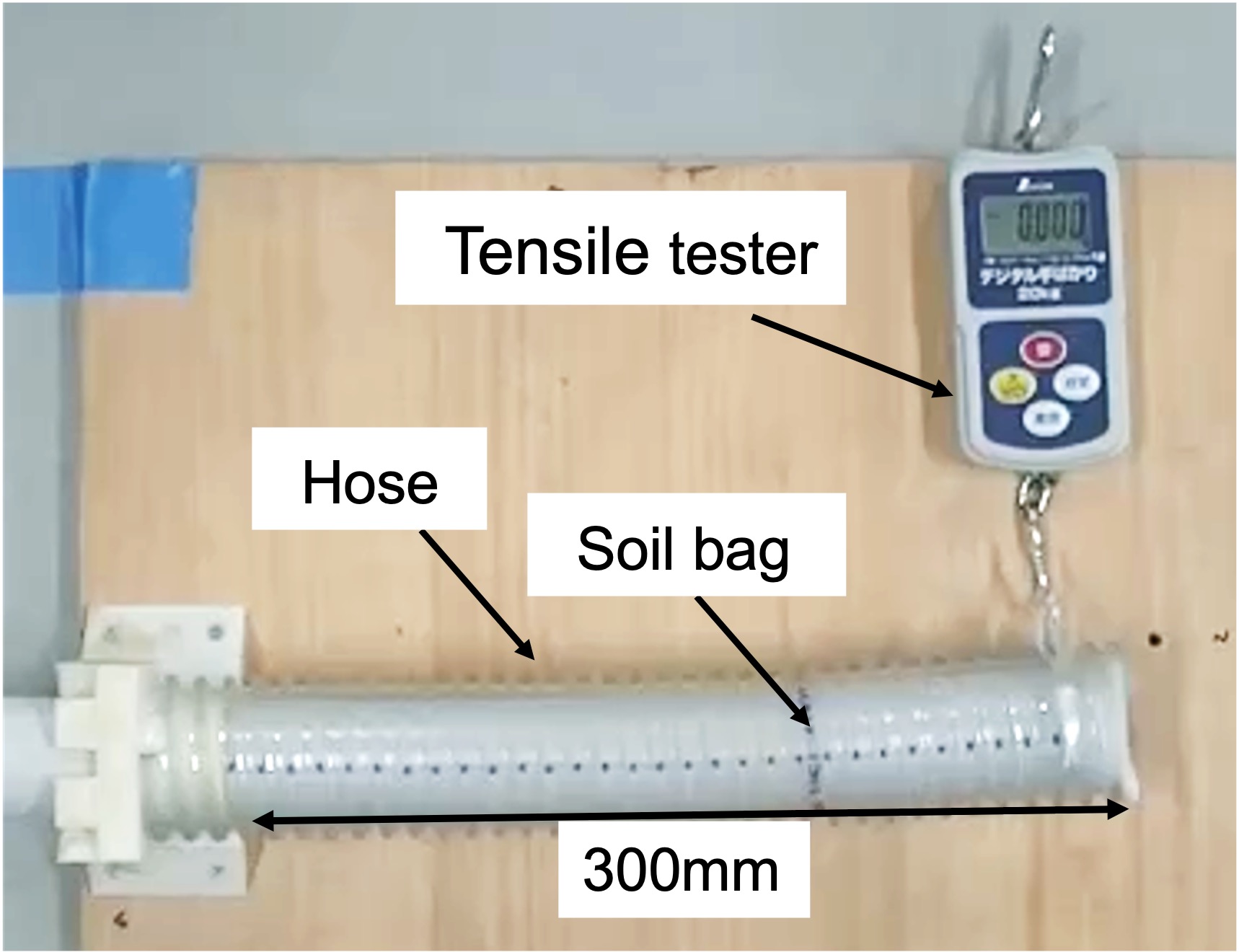}
    \caption{Stiffness measurement experiment environment}\label{fig:mage_souchi}
\end{figure}
\begin{figure}[t]
    \centering
    \includegraphics[width=0.9\columnwidth]{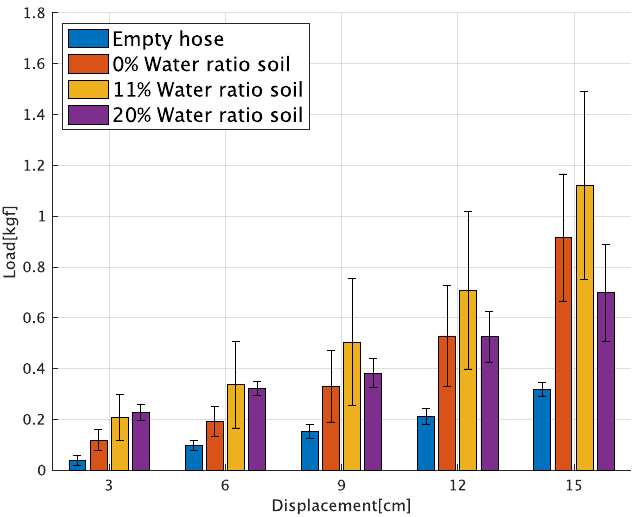}
    \caption{Result of stiffness measurement experiment(Bar graphs are the average of all experiments, and error bars are the standard deviation)}\label{fig:all}
\end{figure}
\subsection{Evaluation of Robot Rigidity by Acquired Soil}
\label{Sec:gosei_experiment}
Next, we experimentally verify the robot's rigidity change by using soil as a structure.
This study measured a load-displacement relationship using a bag filled with soil if the robot had eaten the soil.
We used a flexible hose attached to a fixed part (the same material used for the developed machine), and bags filled with three soil types.
The bags were cylindrical nylon fabric bags 40 mm in diameter and 600 mm long.
We prepared three different types of fresh soil with moisture contents of 0, 11 and 20\%.
We twisted the end of the soil bag, squeezed the entire bag, and applied external pressure to the soil.
The load-displacement relationship was measured from the video images taken when a tensile load was applied to a section 300 mm from the fixed section, as shown in Fig.~\ref{fig:mage_souchi}.
The above measurements were made five times each for four different cases: when the flexible hose was used alone and when the flexible hose was filled with each of the three types of soil bags.

Fig.~\ref{fig:all} shows the results of the stiffness experiments.
As shown in Fig.~\ref{fig:all}, the stiffness increased when the robot was filled with soil, and the 11\% hydrated soil had the highest stiffness.
These results indicate that it is possible to change the properties of the robot, such as rigidity, depending on the soil acquired in the field.
On the other hand, when the robot was filled with soil, its rigidity decreased in the second and subsequent experiments compared to the first experiment.
The reason for this result was that the bending of the bag caused particle slippage and cracking of the bag's surface.
This result should be taken into consideration when using this soil as a structure.
In this experiment, the bag was filled with soil, and the entire bag was squeezed by applying torsion to the end of the bag to apply compression to the soil.
This method was inspired by a previous study in which a bag filled with powder and grains was compressed and then connected at the tip and root with wires to generate tension on the structure's surface, thus gaining greater rigidity\cite{jamming_arm}.
In future work, we plan to increase the number of pouch structures to increase the stiffness of the entire arm and achieve arm-like motion by controlling the tension of the pouch like a wire.

\section{Discussion of Experimental Results Based on Implicit-Explicit Control}
In this section, we discuss the experimental results of two GREEMAs from the viewpoint of Implicit-Explicit control and describe the design guidelines for GREEMAs.
First, we discuss the results of the Mizu-Kurai experiment described in Section 2.
As described in Subsection 2.3, the robot without SAP could not swim properly.
This result can be interpreted as a weakening of implicit control due to the inadequate body structure for swimming, which prevents the robot from taking advantage of interactions from the surrounding environment (water).
In contrast, the robot with SAP swam more efficiently than the robot without SAP.
In other words, the use of SAP to passively take in water from the environment can be interpreted as passively changing the robot's body shape and rigidity without increasing the robot's explicit control (explicit control), resulting in greater implicit control and swimming.
These considerations suggest that the following two points are important in the design of the proposed GREEMA: (1) eating method of environmental objects with small explicit control, and (2) body modification and explicit control design using environmental objects such that the interaction between the body and the environment is implicit control.

Next, we discuss the results of Tsuchi-Kurai.
We found that Tsuchi-Kurai is similar to Mizu-Kurai in that it changes the physical characteristics of the body by taking sediment into the body with a simple explicit control.
In addition, the friction between the bag-shaped fabric and the sediment can be regarded as implicit control in the transport of sediment inside the body by retracting the bag structure.
In other words, the interaction between the robot's internal environmental objects (internal environmental objects) and the body must be appropriately designed to meet the robot's control objectives.
In this paper, due to the difficulty of accurately measuring the interaction between the environment and the body in real time, the qualitative effects of implicit control were demonstrated by comparing the experimental results with and without the inclusion of environmental objects.
Future research should include the development of a mathematical model and quantitative performance evaluation to clarify a more theoretical design method for GREEMA.

\section{Conclusion and Future Work}
In this study, we proposed GREEMA, a robot for immediate response to river channel blockage, which dramatically changes its physical characteristics by taking environmental materials into its body to express its functions, and then removes the materials.
We developed a water- and sediment-eating robot and verified its validity through experiments.
We confirmed that the robot, which passively absorbs water by using water-absorbing polymers, can perform swimming functions by implicitly acquiring a flexible body that is effective for swimming.
In the development of a sediment-eating robot, we proposed a method of taking sediment into the body using friction caused by the pulling of cloth and verified the validity of this method through eating experiments.
In addition, we conducted a rigidity verification experiment on a hose that taked sediment into the body and showed that the acquired sediment increased the rigidity of the body and that the rigidity changed depending on the amount of water content.
Finally, based on the results of these two experiments, we concluded that the body and control design of GREEMA should be conducted so that the interaction between the internal environment and the robot's body, as well as the interaction between the body and the external environment after the robot has been captured, are appropriate for the robot's purpose.
The robot developed in this paper will contribute to the development of new methods of drainage and soil excavation that do not rely on the robot's weight.
Future research will focus on the development of robots that can function by eating other environmental objects, such as trees and sand (lunar regolith).

\acknowledgements
This research was supported in part by grants-in-aid for JSPS KAKENHI Grant Number JP22K14277 and JST Moonshot Research and Development Program JPMJMS2032 (Innovation in Construction of Infrastructure with Cooperative AI and Multi Robots Adapting to Various Environments).

{\small
\bibliography{myref_greema}
\bibliographystyle{junsrt}
}

{\bf Supporting Online Material:}
{\small
\begin{enumerate}
\item[[a\textrm{]}] NPO Sediment Disaster Prevention Publicity Center Home Page:\url{https://www.sabopc.or.jp/library/river_blockage/}
\end{enumerate}
}
\begin{profile}
    \Name{Yusuke Tsunoda}
    \Affiliation{Project assistant professor, Department of Mechanical Engineering, Osaka University}
    \Address{2-1, Yamadaoka, Suita, Osaka 565-0871, JAPAN}
    \History{2021- Project assistant professor, Osaka University}
    \Works{
        $\bullet$ Yusuke Tsunoda, Yuichiro Sueoka, Yuto Sato, Koichi Osuka: “Analysis of Local-camera based Shepherding Navigation”, Advanced Robotics, Vol. 32. No. 23, pp. 1217-1228, 2018 (https://doi.org/10.1080/01691864.2018.1539410).\\
        $\bullet$ Yusuke Tsunoda, Yuichiro Sueoka, Koichi Osuka: “Experimental Analysis of Acoustic Field Control-based Robot Navigation”, Journal of Robotics and Mechatronics, Vol. 31, No. 1, pp.110-117, 2019 (doi: 10.20965/jrm.2019.p0110).
    }
    \Membership{$\bullet$ The Japan Society of Mechanical Engineers (JSME).\\
        $\bullet$ The Institute of Systems Control and Information Engineers (ISCIE).\\
        $\bullet$ Society of Instrument and Control Engineers (SICE).}
    \Photo[\epsfig{file=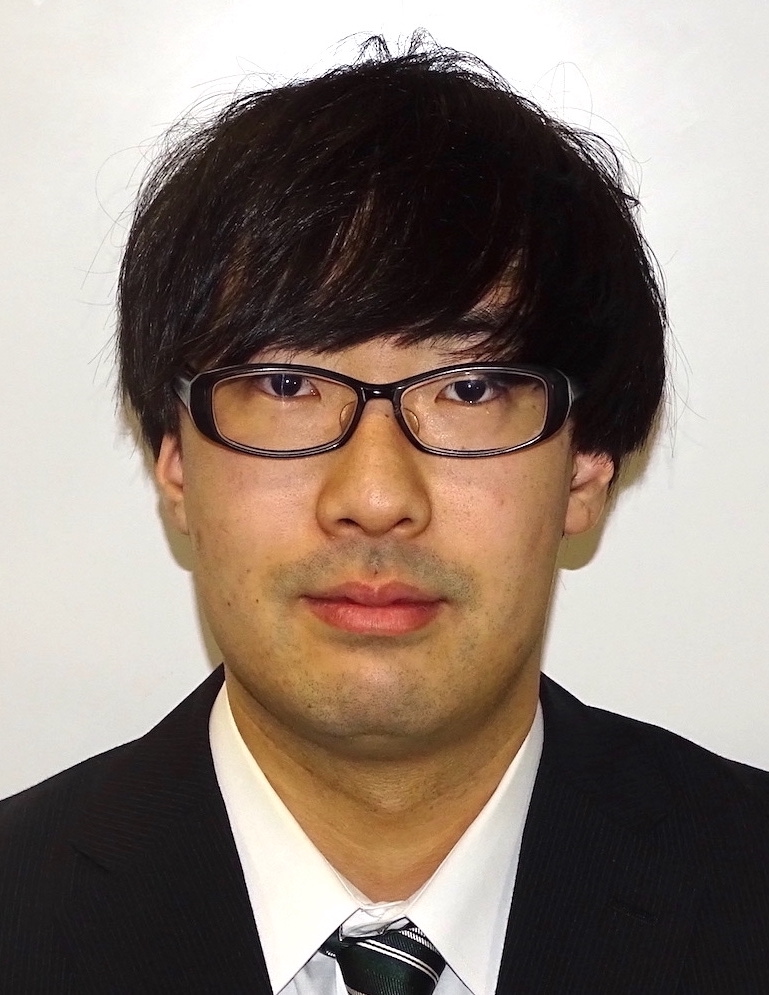,height=93.54pt, width=70.86pt}]
\end{profile}
\begin{profile}
    \Name{Yuya Sato}
    \Affiliation{Master's cource student, Department of Mechanical Engineering, Osaka University}
    \Address{2-1, Yamadaoka, Suita, Osaka 565-0871, JAPAN}
    \History{2022- Master's cource, Osaka University}
    \Photo[\epsfig{file=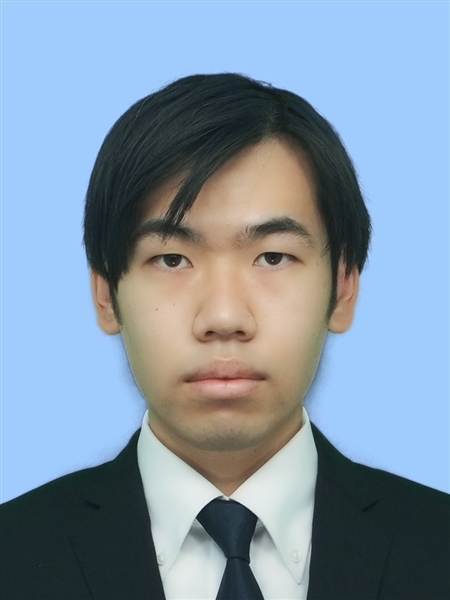,height=93.54pt, width=70.86pt}]
\end{profile}
\begin{profile}
    \Name{Koichi OSUKA}
    \Affiliation{Professor, Department of Mechanical Engineering, Osaka Univ. CREST, Japan Science and Technology Agency }
    \Address{2-1, Yamadaoka, Suita, Osaka, 565-0871, JAPAN}
    \History{1986- Research Associate-Assistant Prof., Osaka Prefecture University\\
        1998- Assistant Professor,Kyoto University\\
        2003- Professor.,Kobe University\\
        2009- Professor:Osaka University
    }
    \Works{$\bullet$ Yuichiro Sueoka, Takuto Kita, Masato Ishikawa, Yasuhiro Sugimoto and Koichi Osuka: “Distributed control of the number of clusters in obstacle collecting by swarm agents”, Nonlinear Theory and Its Applications, Vol.5, No.4, pp.476-486, 2014.\\
        $\bullet$ T. Kinugasa, T. Akagi, T. Haji, K. Yoshida, H. Amano, R. Hayashi, M. Iribe, K. Tokuda, K. Osuka: “Measurement System for Flexed Shape of Flexibly Articulated Mobile Track”, Journal of Intelligent \& Robotic Systems, Vol.75, No.1, pp.87-100, 2014.\\
        $\bullet$ Yuichiro Sueoka, Takamasa Tahara, Masato Ishikawa, Koichi Osuka: “On Statistical Analysis of Object Pattern Formation by Autonomous Transporting Agents”, 2014 Int'l. Symp. on Nonlinear Theory and its Applications (NOLTA2014), pp.854-857, 2014.
    }
    \Membership{$\bullet$ The Philosophical Association of Japan.\\
        $\bullet$ The Institute of Electrical and Electronics Engineers (IEEE).\\
        $\bullet$ The Japan Society of Mechanical Engineers (JSME).\\
        $\bullet$ The Robotics Society of Japan (RSJ).}
    \Photo[\epsfig{file=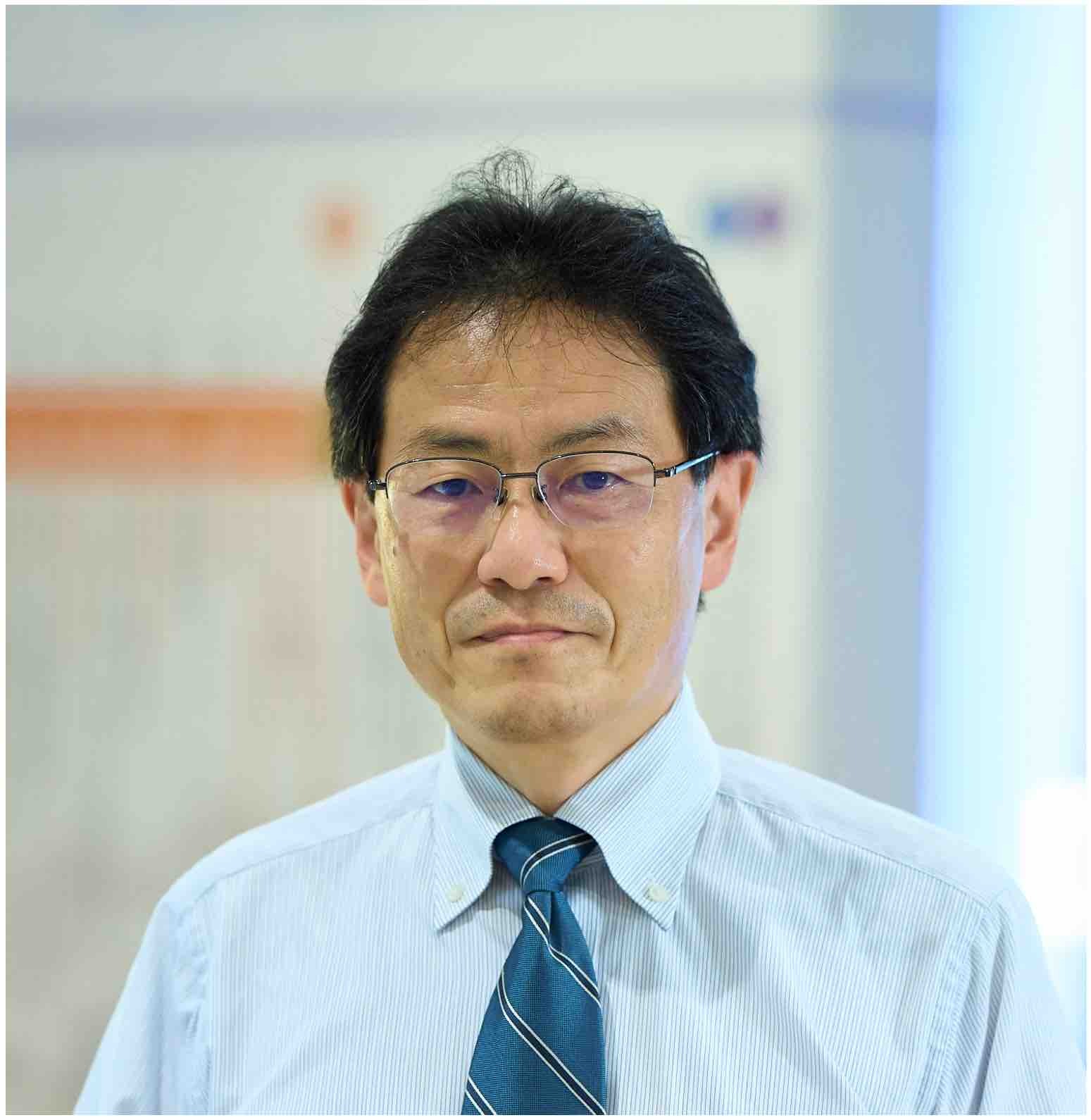,width=70.86pt}]
\end{profile}
\end{document}